\documentclass[acmtog,nonacm]{acmart}
\usepackage{mathtools}
\usepackage{enumitem}
\usepackage{multirow}
\usepackage{tcolorbox}
\usepackage[safe]{tipa}
\setlist[enumerate]{label=(\roman*),leftmargin=*}   

\AtBeginDocument{%
  }

\acmConference[Preprint]{arXiv preprint}{July}{2025}

\begin{document}


\title{Harnessing Diffusion-Yielded Score Priors for Image Restoration}


\author{Xinqi Lin}
\email{linxinqi23@mails.ucas.ac.cn}
\affiliation{
  \institution{Shenzhen Institute of Advanced Technology, CAS and University of Chinese Academy of Sciences}
  \city{Shenzhen}
  \country{China}
}

\author{Fanghua Yu}
\email{fanghuayu96@gmail.com}
\affiliation{
  \institution{Shenzhen Institute of Advanced Technology, CAS}
  \city{Shenzhen}
  \country{China}
}

\author{Jinfan Hu}
\email{jf.hu1@siat.ac.cn}
\affiliation{
  \institution{Shenzhen Institute of Advanced Technology, CAS and University of Chinese Academy of Sciences}
  \city{Shenzhen}
  \country{China}
}

\author{Zhiyuan You}
\email{zhiyuanyou@link.cuhk.edu.hk}
\affiliation{
  \institution{Shenzhen Institute of Advanced Technology, CAS and The Chinese University of Hong Kong}
  \city{Shenzhen}
  \country{China}
}

\author{Wu Shi}
\email{wu.shi@siat.ac.cn}
\affiliation{
  \institution{Shenzhen Institute of Advanced Technology, CAS}
  \city{Shenzhen}
  \country{China}
}

\author{Jimmy S. Ren}
\email{jimmy.sj.ren@gmail.com}
\affiliation{%
 \institution{SenseTime Research and Hong Kong Metropolitan University}
 \city{Hong Kong}
 \country{China}}

\author{Jinjin Gu}
\authornote{Chao Dong and Jinjin Gu are joint corresponding authors.}
\email{jinjin.gu@insait.ai}
\affiliation{%
  \institution{INSAIT, Sofia University}
  \city{Sofia}
  \country{Bulgaria}}

\author{Chao Dong*}
\email{chao.dong@siat.ac.cn}
\affiliation{%
  \institution{Shenzhen Institutes of Advanced Technology, CAS and Shenzhen University of Advanced Technology}
  \city{Shenzhen}
  \country{China}}

\renewcommand{\shortauthors}{Xinqi Lin, Fanghua Yu, Jinfan Hu, Zhiyuan You, Wu Shi, Jimmy S. Ren, Jinjin Gu, Chao Dong}

\begin{teaserfigure}
\centering
  \includegraphics[width=0.95\textwidth]{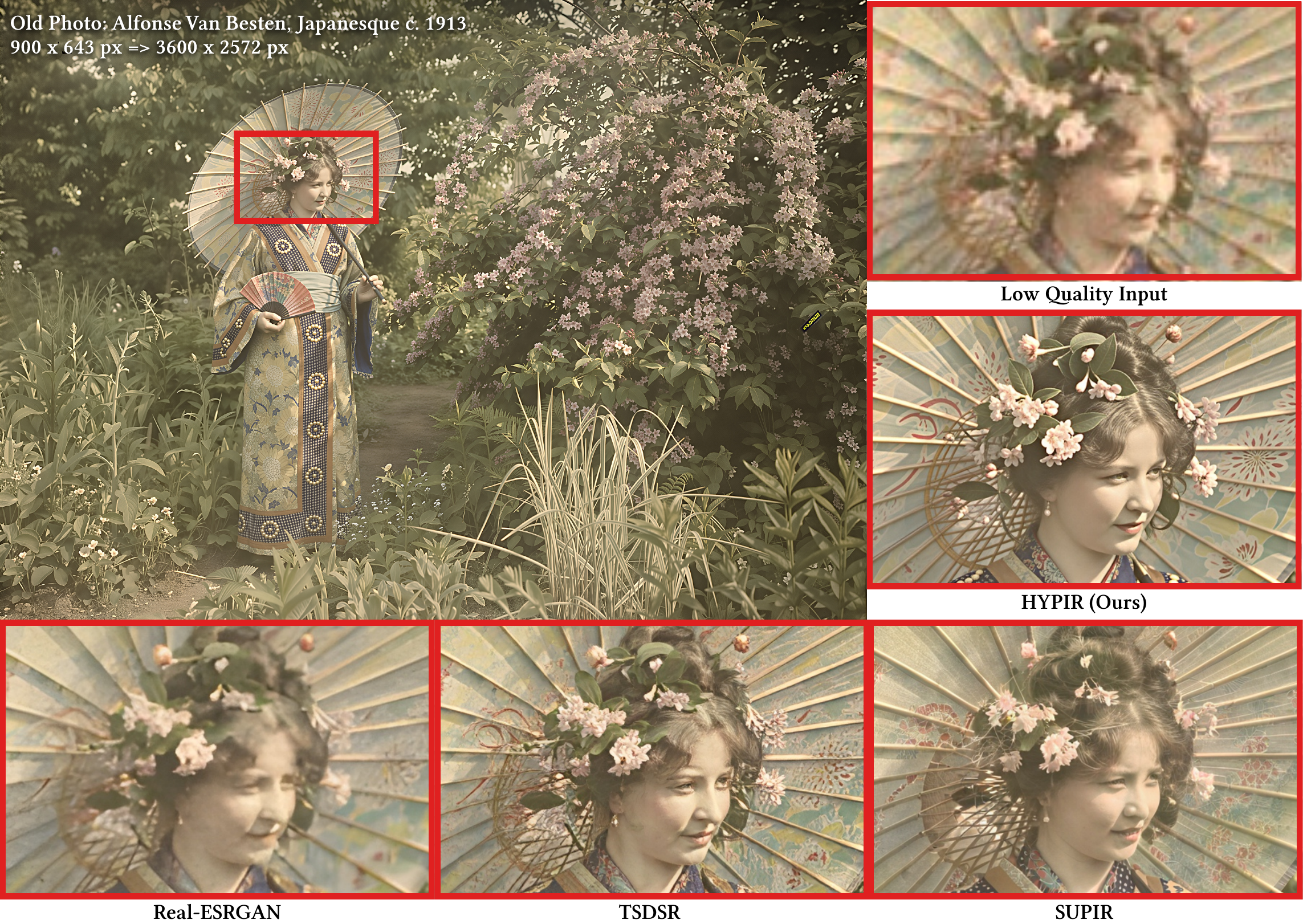}
  \vspace{-2mm}
  \caption{We present HYPIR, a new image restoration framework that leverages pretrained diffusion models as strong generative priors. HYPIR discards the iterative sampling process inherent in diffusion, and instead directly fine-tunes a generative adversarial network. HYPIR delivers superior visual quality compared to state-of-the-art approaches. By eliminating the need for multi-step iterative inference, our method also achieves significant improvements in computational efficiency. Photo Credit: \textit{Japanesque} (Autochrome, ca.\ 1913). Photograph by Alfonse Van Besten, Public Domain. Source: Wikimedia Commons.}
  \Description{We present HYPIR, a new image restoration framework that leverages pretrained diffusion models as strong generative priors. HYPIR discards the iterative sampling process inherent in diffusion-based methods, and instead directly fine-tunes a generative adversarial network. HYPIR delivers superior visual quality compared to state-of-the-art approaches. By eliminating the need for multi-step iterative inference, our method also achieves significant improvements in computational efficiency. Photo Credit: \textit{Japanesque} (Autochrome, ca.\ 1913). Photograph by Alfonse Van Besten, Public Domain. Source: Wikimedia Commons.}
  \label{fig:teaser}
\end{teaserfigure}


\begin{abstract}
Deep image restoration models aim to learn a mapping from degraded image space to natural image space.
However, they face several critical challenges: removing degradation, generating realistic details, and ensuring pixel-level consistency.
Over time, three major classes of methods have emerged, including MSE-based, GAN-based, and diffusion-based methods.
However, they fail to achieve a good balance between restoration quality, fidelity, and speed.
We propose a novel method, HYPIR, to address these challenges.
Our solution pipeline is straightforward: it involves initializing the image restoration model with a pre-trained diffusion model and then fine-tuning it with adversarial training. This approach does not rely on diffusion loss, iterative sampling, or additional adapters.
We theoretically demonstrate that initializing adversarial training from a pre-trained diffusion model positions the initial restoration model very close to the natural image distribution. Consequently, this initialization improves numerical stability, avoids mode collapse, and substantially accelerates the convergence of adversarial training.
Moreover, HYPIR inherits the capabilities of diffusion models with rich user control, enabling text-guided restoration and adjustable texture richness.
Requiring only a single forward pass, it achieves faster convergence and inference speed than diffusion-based methods.
Extensive experiments show that HYPIR outperforms previous state-of-the-art methods, achieving efficient and high-quality image restoration.

\end{abstract}




\maketitle

\section{Introduction}
In the era of deep learning, the essence of image restoration algorithms lies in learning an effective mapping from the degraded image distribution back to the natural image distribution.
There are three major challenges: 
(1) Removing degradation from the input image.
(2) Generating outputs that faithfully adhere to the natural image distribution.
(3) Ensuring pixel-level consistency and alignment between the degraded and restored images.
The historical evolution of image restoration algorithms can be viewed as a continuous process of overcoming these key challenges.

\begin{figure}
    \centering
    \includegraphics[width=\linewidth]{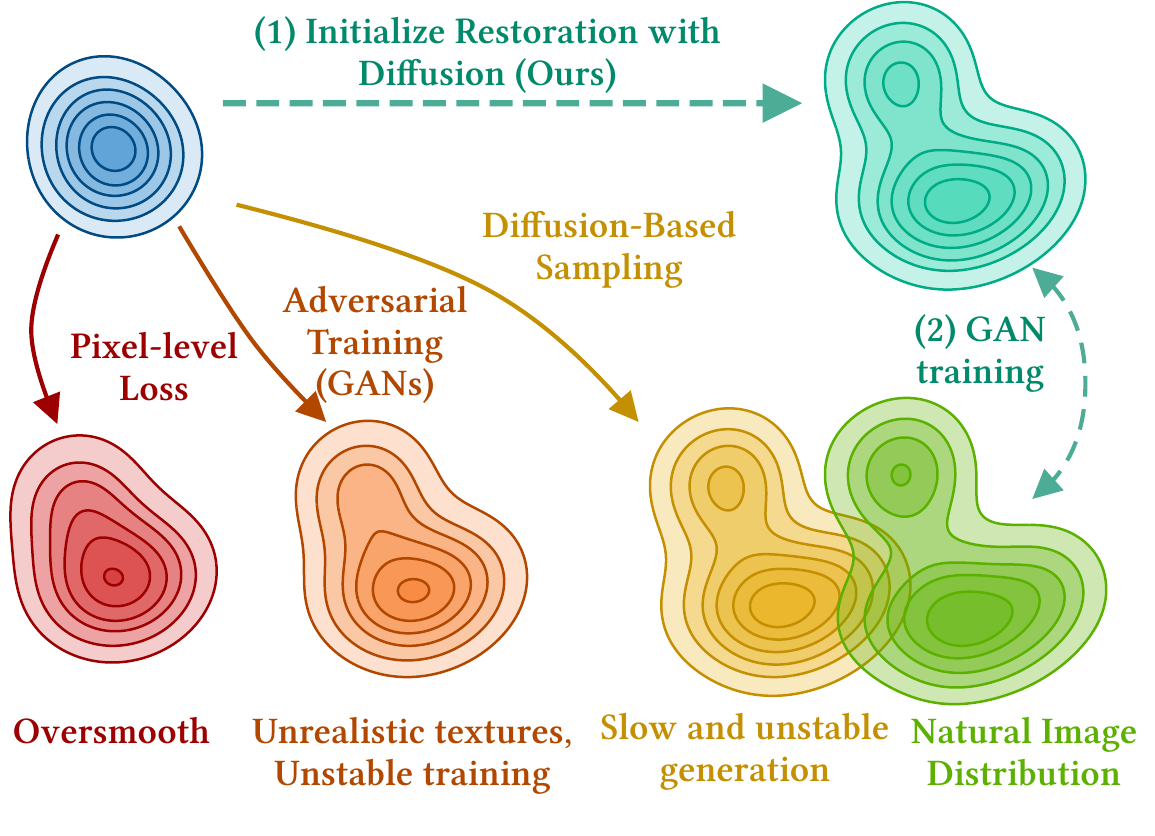}
    \caption{Existing pixel-level loss, adversarial training, and diffusion-based image restoration methods struggle with over-smoothness, unrealistic textures, and slow, unstable generation. Our approach leverages diffusion initialization followed by GAN training, effectively balancing realism and efficiency.}
    \Description{Existing pixel-level loss, adversarial training, and diffusion-based image restoration methods struggle with over-smoothness, unrealistic textures, and slow, unstable generation. Our approach leverages diffusion initialization followed by GAN training, effectively balancing realism and efficiency.}
    \label{fig:teaser-2}
\end{figure}

Since 2014, convolutional neural networks \cite{dong2015image,dong2016accelerating} employing pixel-level loss functions have marked the beginning of the deep-learning era in image restoration.
They have effectively addressed the first and third challenges, achieving impressive quantitative restoration performance as indicated by metrics such as PSNR and SSIM.
However, these methods typically produce overly smooth images that lack realistic details \cite{blau2018perception,jinjin2020pipal,gu2020image}.
To address the second challenge, generative adversarial network (GAN)-based restoration models emerged around 2017 \cite{ledig2017photo,wang2018esrgan}.
These approaches employ perceptual loss along with adversarial training to produce restoration results that are better aligned with human visual perception.
However, training such GAN-based models has never been easy.
Existing methods struggle to capture the vast diversity of the natural image space, typically generating only a limited set of common/unrealistic textures rather than the complete spectrum of image details \cite{wang2018esrgan,jinjin2020pipal}.
In recent years, diffusion-based image restoration algorithms leveraging large-scale pretrained text-to-image models have successfully overcome previous bottlenecks, producing highly realistic images \cite{lin2024diffbir,yu2024scaling,yu2025unicon}.
However, the inherent multi-step iterative nature of diffusion models leads to critical drawbacks, including prolonged training time, slow inference speed, and unstable generation outcomes.
While some recent diffusion distillation methods can reduce the number of diffusion steps, they still cannot fundamentally overcome these intrinsic limitations of diffusion models, often struggling to balance generation quality with computational efficiency.
An illustrative comparison of these three classes of methods is presented in \figurename~\ref{fig:teaser-2}.

In this work, we introduce a new methodology to construct image restoration models, specifically designed to address these three challenges.
Our approach is simple: it leverages a pretrained diffusion model for effective parameter initialization, followed by lightweight adversarial fine-tuning to adapt it specifically for image restoration tasks.
Notably, our method does NOT rely on diffusion teachers, control adapters, and iterative refinement steps.
As a result, both training and inference speeds are more than an order of magnitude faster than diffusion-based methods, with even superior performance.
Since it \textbf{H}arnesses diffusion-\textbf{Y}ield Score \textbf{P}rior for \textbf{I}mage \textbf{R}estoration, we name the proposed method \textbf{HYPIR} (\textipa{/""haI.p@r/}).

Why does such a simple approach work so well?
In essence, two observations are sufficient.
First, we show that image restoration amounts to estimating the score, namely the gradient of the log-density of degraded images, which points along the quickest route back to the natural image distribution.
Because diffusion models are trained to learn exactly such score fields across noise levels, the prior they internalize is already a close match to the ideal restoration operator.
Second, initializing the restoration network with diffusion weights places it near the natural image distribution, so that adversarial gradients remain small and numerically stable.
This favorable initialization spans almost all modes of the data, guards against collapse, and drives much faster convergence to high-fidelity results than training from scratch.
Together, these insights lead to a conclusion: diffusion pre-training provides a near-optimal initialization for image restoration, while lightweight adversarial fine-tuning aligns the model closely with the real data distribution.
We underpin both observations with heuristic proofs.

Despite its simplicity, this strategy has been overlooked because of two persistent mindsets.
First, diffusion models are almost always paired with diffusion-style samplers, so previous works remain trapped in iterative pipelines and distillation schemes.
Second, the research field widely assumes that diffusion models consistently outperform GANs, making pure adversarial training look unpromising for cutting-edge results.
Yet the evidence is counter-intuitive: with a good enough initialization, the GAN objective outpaces the diffusion objective in convergence speed, training stability, output fidelity, and inference effectiveness.
A well-trained diffusion model supplies that initialization.

HYPIR delivers compelling practical benefits.
\textbf{Training-wise}, it converges within only tens of thousands of training steps.
Because it does not require any auxiliary ControlNet, its memory footprint remains low enough to accommodate the latest and largest diffusion models as priors.
\textbf{Inference-wise}, a single forward pass is enough, eliminating the costly diffusion sampling loop.
\textbf{Performance-wise}, HYPIR consistently surpasses the state-of-the-art restoration networks, while also enabling rich user control: it can follow text prompts, adjust texture richness, and balance generative enhancement against fidelity.
\figurename~\ref{fig:teaser} shows one of the surprising effects of HYPIR, and highlights the effectiveness of our method.

\section{Related Work}

\subsection{Image Restoration}
Image restoration aims to recover high-quality images from their degraded counterparts \cite{liang2021swinir,jinjin2020pipal,zhang2017learning,zhang2022accurate,zhang2019residual}.
%
%
Image restoration has evolved significantly, progressing from traditional filtering and optimization-based methods \cite{rudin1992nonlinear,tomasi1998bilateral,dabov2007image} to advanced deep learning approaches \cite{dong2015image,chen2023dual,gu2019blind}, and from handling single-degradation scenarios \cite{liang2021swinir,chen2023recursive} to addressing multiple degradations simultaneously \cite{wang2021real,zhang2021designing,chen2023masked}.
Recent state-of-the-art image restoration models often train on diverse mixed degradations \cite{zhang2021designing,wang2021real,lin2024diffbir,yu2024scaling,kong2022reflash}, enabling a single model to generalize across various real-world degradation scenarios.
These models are commonly referred to as real-world image restoration models.
In this work, we similarly address the challenge of constructing a unified, advanced image restoration model capable of managing diverse and complex degradation conditions.

\subsection{Generative Prior}
Image restoration faces several practical challenges, among which producing photo-realistic results remains particularly challenging.
For severely degraded images, pixel-wise loss functions often result in overly smooth outputs, deviating from the natural image distribution \cite{blau2018perception,ledig2017photo}.
Generative models, or generative priors, are widely recognized as effective solutions to this problem.
By explicitly modeling and generating images that align with natural image distributions, these models can ensure visually realistic outputs.
Two paradigms have emerged to address this issue.

\subsubsection{Adversarial Training}
Generative adversarial networks (GANs) \cite{goodfellow2014generative} were among the earliest generative methods proven effective for natural image generation \cite{radford2015unsupervised,karras2017progressive,karras2019style}.
Adversarial training consists of a generator and a discriminator: the generator aims to produce images indistinguishable from real ones, while the discriminator attempts to differentiate between generated and real images.
Through alternating training, the generator ideally learns to fool the discriminator.
Theoretically, adversarial training can be viewed as minimizing certain divergences between the generated and real distributions, such as the Kullback–Leibler (KL) \cite{nowozin2016f} or Wasserstein distances \cite{arjovsky2017wasserstein}.
Adversarial generative models have been extensively applied in image restoration tasks, including GAN inversion \cite{pan2021exploiting,gu2020image,bau2020semantic}, GAN encoders \cite{chan2021glean,zhu2022disentangled}, direct adversarial training \cite{ledig2017photo,wang2018esrgan}, or utilizing GAN as core restoration modules \cite{yang2021gan,wang2021towards}.
However, adversarial training suffers from notable challenges, such as mode collapse, training instability, scaling difficulties, and suboptimal generative quality \cite{salimans2016improved,kodali2017convergence,odena2017conditional,lucic2018gans,mescheder2018training}.
These issues have gradually diminished the prominence of GAN-based methods in recent state-of-the-art image restoration research.

\begin{figure*}
    \centering
    \includegraphics[width=\linewidth]{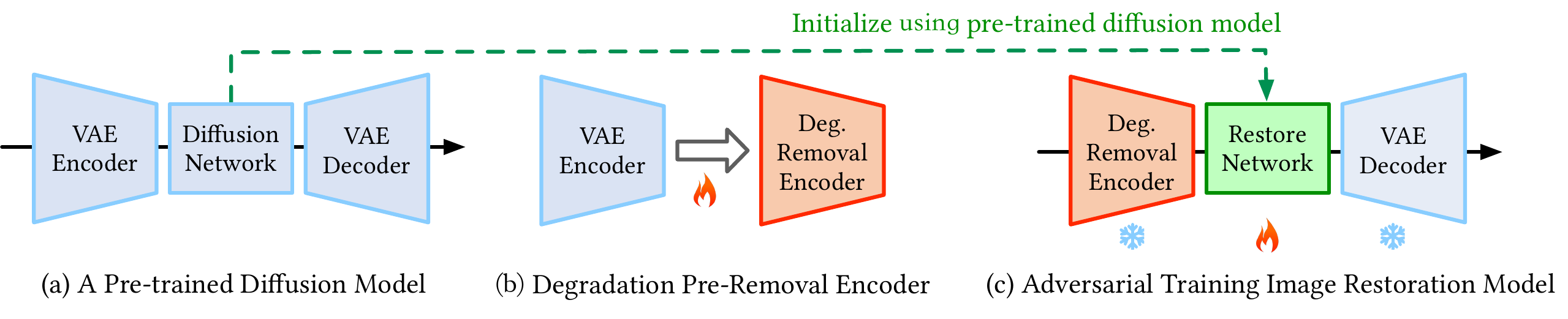}
    \caption{Illustration of our proposed image restoration pipeline. (a) We start with a pre-trained diffusion model. (b) The VAE encoder is fine-tuned specifically for degradation pre-removal, enhancing robustness against severe image degradation. (c) Subsequently, the degradation-aware encoder and pre-trained decoder initialize an adversarially-trained image restoration model, where only the ``Restore Network'' is optimized during this stage.}
    \label{fig:method-1}
\end{figure*}

\subsubsection{Diffusion-based Models}
Diffusion models have emerged recently as a powerful class of generative techniques, quickly becoming mainstream in natural image generation and restoration \cite{ho2020denoising,song2020denoising,dhariwal2021diffusion,rombach2022high,peebles2023scalable}.
These models introduce stochastic noise incrementally (forward diffusion process) and subsequently learn to remove it step by step (reverse denoising process), thereby effectively modeling the image distribution.
Fundamentally, diffusion models optimize an objective related to the score function -- the gradient of the log probability density of the data distribution \cite{song2019generative,song2020denoising}.
As a powerful generative model, diffusion models have also been widely applied to image processing and restoration \cite{saharia2022image,kawar2022denoising,lin2024diffbir,yu2024scaling,yang2024pixel,tao2024overcoming,wang2024exploiting}.
This is primarily achieved by guiding or controlling the diffusion process to generate images that share content with degraded inputs but exclude the degradation itself \cite{zhang2023adding,yu2025unicon}.
Compared with GAN-based models, diffusion models offer enhanced training stability, superior image quality, and improved mode coverage, significantly alleviating mode collapse issues.
However, diffusion models typically require multiple iterative denoising steps (often 50, 100, or even more), resulting in computationally intensive and time-consuming inference.
For instance, state-of-the-art diffusion-based image restoration methods, despite producing groundbreaking results, still need approximately 20 seconds to process a 1080p image, and nearly 3 minutes for 4K images, even under optimized conditions \cite{yu2024scaling,yu2025unicon}.

\subsection{Larger GAN or Faster Diffusion?}
GAN-based methods provide end-to-end, rapid inference but typically yield suboptimal image quality.
In contrast, diffusion models achieve remarkable image quality but involve multi-step inference processes that are computationally expensive.
Recent research efforts have begun exploring the combination of GANs and diffusion models, aiming to leverage their respective strengths to achieve both fast inference and high-quality restoration.
Two distinct approaches have emerged in recent literature: one involves training larger, more powerful GAN models \cite{kang2023scaling}.
However, GAN-based restoration methods continue to face significant mode collapse issues and struggle to produce consistently high-quality results.
The other approach involves distilling diffusion models into smaller diffusion models \cite{salimans2022progressive,sauer2024adversarial,luo2023diff,yin2024one,kang2024distilling} or drastically reducing their inference steps \cite{song2023consistency,luo2023latent}, inevitably leading to a substantial drop in generation quality.
These few-step diffusion or distilled models have also been extensively explored for image restoration or processing tasks \cite{xie2024addsr,wang2024sinsr,yue2023resshift,wu2024one,yue2024arbitrary}.
However, balancing computational efficiency and performance remains particularly challenging within the diffusion model framework.
Currently, neither direction has produced fully satisfactory solutions.

\section{Methodology}
Our approach is remarkably simple.
%
%
The core idea is: \textbf{using a pre-trained diffusion model as initialization, then applying adversarial fine-tuning for restoration}. 
Our method pipeline is shown in \figurename~\ref{fig:method-1}.
Table~\ref{tab:notations} lists all mathematical notations used in this paper along with their explanations.

Let $I$ be a natural clean image, and $I_{\mathrm{deg}}$ be its degraded counterpart.
To process images in the latent space, we utilize the variational encoder $\mathcal{V}_E$ and decoder $\mathcal{V}_D$, inherited from the diffusion model.
Denote $\mathbf{x} = \mathcal{V}_E(I)$ as the latent encoding of the clean natural image.
We assume $\mathbf{x} \sim p_{\text{data}}$, where $p_{\text{data}}$ represents the latent prior distribution of natural images.
In the restoration problem, the degraded image $I_{\mathrm{deg}}$ is mapped into the same latent space.
The encoder $\mathcal{V}_E$ often partially addresses degradation removal, a step known as degradation pre-removal \cite{lin2024diffbir,yu2024scaling}.
To enhance the encoder's robustness to degradation, we fine-tune the encoder $\mathcal{V}_{E_R}$ by minimizing the following objective function, as shown in \figurename~\ref{fig:method-1} (b):
$$
\mathcal{L}_{E} = \|\mathcal{V}_D(\mathcal{V}_{E_R}(I_{\mathrm{deg}})) - \mathcal{V}_D(\mathcal{V}_{E_R}(I_{\mathrm{GT}}))\|_2^2,
$$
where $\mathcal{V}_{E_R}$ is the fine-tuned encoder, $\mathcal{V}_D$ is the fixed decoder.
%
%
The fine-tuned encoder $\mathcal{V}_{E_R}$ maps the degraded image $I_{\mathrm{deg}}$ into a latent observation $\mathbf{y}=\mathcal{V}_{E_R}(I_{\mathrm{deg}})$.
The degraded latent observation $\mathbf{y}$ can be modeled as:
$$
\mathbf{y} = \mathbf{k}_{\mathrm{deg}} * \mathbf{x} + \boldsymbol{\varepsilon},\quad \boldsymbol{\varepsilon}\sim\mathcal{N}(\mathbf{0}, \eta^2\mathbf{I}),
$$
where $\mathbf{k}_{\mathrm{deg}}$ denotes a degradation kernel applied in the latent space, $*$ indicates spatial convolution operating directly on latent representations, and $\boldsymbol{\varepsilon}$ represents additive noise.
This process generates the degraded latent distribution $p_{\mathbf{y}}$.
Due to degradation pre-removal, the latent $\mathbf{y}$ after encoding typically exhibits simpler degradation patterns, characterized mainly by smoothing and adding noise, which can be analogous to the effects of directly applying an MSE-based restoration method.
The above observation model succinctly captures the relationship between the clean latent $\mathbf{x}$ and the degraded latent $\mathbf{y}$.

Consider a differentiable image processing neural network $\mathcal{U}_\theta$, parameterized by $\theta$.
Our goal is to optimize parameters $\theta$ so that the push-forward distribution
\[
p_{\theta}\;=\;\mathcal{U}_{\theta}\,\sharp\,p_{y}
\;\;\bigl(\text{i.e.\ }\hat{\mathbf{x}}\sim p_{\theta}\iff\hat{\mathbf{x}}=\mathcal{U}_{\theta}(\mathbf{y}),\;\mathbf{y}\sim p_{y}\bigr),
\]
matches $p_{\mathrm{data}}$ as closely as possible when driven by degraded inputs $\mathbf{y} \sim p_{y}$.
A prevalent approach that achieves this goal is through adversarial training, optimizing the statistical distance (e.g., Jensen-Shannon or Wasserstein distances) between distributions $p_{\theta}$ and $p_{\mathrm{data}}$ \cite{nowozin2016f}.
A discriminator $D_{\phi}$, parameterized by $\phi$, is trained concurrently to differentiate whether a given sample originates from the clean distribution $\mathbf{x}\sim p_{\mathrm{data}}$ or from the restoration network's output $\hat{\mathbf{x}}\sim p_{\theta}$.
This adversarial training objective reads
\begin{equation}
  \min_{\theta}\;\max_{\phi}\;
\Bigl[
\mathbb{E}_{\mathbf{x}\sim p_{\mathrm{data}}} 
  \bigl[\log D_\phi(\mathbf{x})\bigr]
+
\mathbb{E}_{\mathbf{y}\sim p_{y}} 
  \bigl[\log\bigl(1-D_\phi(\mathcal{U}_\theta(\mathbf{y}))\bigr)\bigr]
\Bigr].
\label{eq:cgan}
\end{equation}

For image restoration, it is beneficial to exploit the pixel-wise correspondence between $\mathbf{y}$ and ground-truth $\mathbf{x}$ \cite{ledig2017photo,wang2018esrgan}.
One can augment Equation~\eqref{eq:cgan} with a reconstruction (content) penalty that penalizes deviations of the generator's output from the target image:
\begin{equation}
  \min_{\theta}\;\max_{\phi}\;
\mathcal{L}_{\mathrm{adv}}(\theta,\phi)
\;+\;
\lambda_{\mathrm{rec}}\,
\mathbb{E}_{(\mathbf{y},\mathbf{x})\sim p_{\mathrm{data}}}
 \bigl[\,
\mathrm{Recon}(\mathcal{U}_\theta(\mathbf{y}),\mathbf{x})
\bigr],
\label{eq:total}
\end{equation}
where $\mathcal{L}_{\mathrm{adv}}$ denotes the bracketed term in Equation~\eqref{eq:cgan} and $\lambda_{\mathrm{rec}} > 0$ controls the trade-off between perceptual realism and fidelity to the ground truth.
In practice, we adopt MSE and LPIPS \cite{zhang2018unreasonable} as the reconstruction terms in Equation~\eqref{eq:total}, combining pixel-level accuracy with perceptual similarity.
This is a common training method for GAN-based image restoration models \cite{wang2018esrgan}.

Although GAN-based approaches are designed to generate realistic textures, they traditionally suffer from significant drawbacks, such as training instability and mode collapse.
They fail to fully capture the diversity of the natural image space, often producing only a limited set of common textures.
Consequently, the generated images tend to exhibit uniform patterns, lacking realism and sensitivity to image-specific content.

Our approach makes a crucial modification to the GAN framework: \textbf{we initialize adversarial training from a well-trained diffusion generative model}.
Specifically, suppose a diffusion generative model employs the same neural network $\mathcal U$ as its score function $\mathcal S = \mathcal U_{\theta_\mathrm{Diff}}$, where $\theta_\mathrm{Diff}$ denotes the pretrained parameters of the diffusion model.
Our adversarial training directly begins from these pretrained parameters.
As illustrated in \figurename~\ref{fig:method-1} (c), we inherit both the network architecture from the diffusion model and the previously fine-tuned degradation-removal encoder, thus forming a complete image-to-image restoration pipeline.
During restoration training, we fine-tune only the intermediate U-Net $\mathcal{U}$ while keeping the encoder and decoder fixed.
Due to the large parameter count of advanced diffusion models, we utilize the LoRA \cite{hu2022lora} fine-tuning technique to substantially reduce the number of trainable parameters and accelerate training.

\section{Discussion}
Above, we introduce a simple framework, but the underlying insights are profound.
In this section, we present an analysis of these insights and explore why our straightforward approach could achieve remarkable effectiveness.

\subsection{Why Does Such a Simple Approach Work So Well?}
\label{sec:why_work}
Traditionally, GANs and diffusion models have been considered separate research directions due to their fundamentally different training objectives.
However, upon closer inspection, a pretrained diffusion model is already remarkably close to an ideal image restoration model, and minimal adversarial fine-tuning can transform it into an excellent end-to-end restoration network.
To understand the underlying reasons, we need to address two critical questions succinctly:
(1) Why is a pretrained diffusion model inherently suitable for image restoration?
(2) Why does using this pretrained diffusion model as initialization substantially improve adversarial training for restoration?
We discuss these points in detail next.

\begin{tcolorbox}[colframe=black!50,colback=yellow!10,boxrule=0.5pt, boxsep=1pt]
(1) Why is a pretrained diffusion model inherently suitable for image restoration?
\end{tcolorbox}

Diffusion models have emerged as powerful generative approaches that learn a rich prior over clean image distributions.
We first explain why such pretrained diffusion priors are naturally well-suited for image restoration tasks, even without a diffusion-like sampling procedure.
Our discussion proceeds from the standard forward–reverse stochastic differential equation (SDE) framework to the posterior SDE for conditional sampling, and then highlights the practical challenge of incorporating the observation likelihood.
Our method proposes an approximation strategy that bypasses the intractable terms and exploits the denoising ability of the pre-trained diffusion model.

A diffusion model typically starts with a clean image $\mathbf{x}_0\sim p_{\mathrm{data}}(\mathbf{x}_0)$, then gradually corrupts it via the forward SDE (Equation~\eqref{eq:forwardSDE}):
\begin{equation}
    \begin{aligned}
d\mathbf{x}_t
&=
-\frac{\beta(t)}{2}\,\mathbf{x}_t\,dt \;+\;\sqrt{\beta(t)}\,dw,
\end{aligned}
\label{eq:forwardSDE}
\end{equation}
where $t\in[0,T]$ is the continuous time index, $\beta(t)$ is a noise schedule, and $w$ denotes standard Brownian motion. Let $\mathbf{x}_t$ be a noisy version of $\mathbf{x}_0$ at time $t$. At time $T$, $\mathbf{x}_T$ becomes nearly pure noise. To invert this corruption, one uses the reverse SDE (Equation~\eqref{eq:reverseSDE}:
\begin{equation}
    \begin{aligned}
d\mathbf{x}_t 
&=
\Bigl[
-\tfrac{\beta(t)}{2}\,\mathbf{x}_t
-\beta(t)\,\nabla_{\mathbf{x}_t}\log\,p_t(\mathbf{x}_t)
\Bigr]\,dt
\;+\;\sqrt{\beta(t)}\,d\bar{w},
\end{aligned}
\label{eq:reverseSDE}
\end{equation}
where $p_t(\mathbf{x}_t) = (p_\mathrm{data}*k_\sigma(t))(\mathbf{x}_t)$ denotes the distribution of $\mathbf{x}_t$ at time $t$ with smooth noise $k_\sigma(t)$, and $d\bar{w}$ is a Brownian motion in the reverse-time direction.
In practice, a neural network $s_\theta(\mathbf{x}_t,t)\approx \nabla_{\mathbf{x}_t}\log\,(p_\mathrm{data}*k_\sigma(t))(\mathbf{x}_t)$ is learned to serve as the ``score'' or ``denoising'' function \cite{song2019generative}.

For image restoration, we are given a degraded observation $\mathbf{y}$ and aim to sample $\mathbf{x}_0$ from the posterior $p(\mathbf{x}_0|\mathbf{y})$. In principle, one can define a posterior SDE that integrates both the diffusion prior and the data likelihood, leading to:
\begin{align}
    d\mathbf{x}_t
=&\Bigl[
-\tfrac{\beta(t)}{2}\,\mathbf{x}_t
-\beta(t)\,\nabla_{\mathbf{x}_t}\log p_t(\mathbf{x}_t\mid\mathbf{y})
\Bigr]\,dt\notag\
+\sqrt{\beta(t)}\,d\bar{w},
\\
=&\Bigl[
-\tfrac{\beta(t)}{2}\,\mathbf{x}_t
-\beta(t)\Big(\nabla_{\mathbf{x}_t}\log p_t(\mathbf{x}_t)
+\nabla_{\mathbf{x}_t}\log p_t(\mathbf{y}\mid \mathbf{x}_t)
\Big)\Bigr]\,dt\notag\\&
+\sqrt{\beta(t)}\,d\bar{w}, 
\label{eq:Bayes}
\end{align}  
where $p_t(\mathbf{x}_t|\mathbf{y})\propto p_t(\mathbf{x}_t)\,p_t(\mathbf{y}\mid \mathbf{x}_t)$.
However, the third gradient term
$\nabla_{\mathbf{x}_t}\log p_t(\mathbf{y} \mid \mathbf{x}_t)$
is intractable or highly expensive to approximate. Existing works often rely on heuristic approximations, which can result in suboptimal or inefficient sampling \cite{chung2022diffusion,xu2025rethinking}.
In contrast, we propose to discard the term $\nabla_{\mathbf{x}_t}\log p_t(\mathbf{y}\mid \mathbf{x}_t)$ in Equation~\eqref{eq:Bayes} in order to use the unconditional reverse diffusion dynamics.
Although this seems to lose the explicit dependence on $\mathbf{y}$, we can re-inject the observation later in the following manner:

\begin{enumerate}
    \item \emph{Inject Observation at an Intermediate Time}: We run the forward process (or consider a corrupted sample) until an intermediate time $t$, and then \emph{replace} $\mathbf{x}_t$ with a representation derived from the observed $\mathbf{y}$. Intuitively, after sufficient noise is added, the difference between various degradations becomes less significant, and substituting $\mathbf{y}$ (or its encoded latent) is a reasonable approximation of the posterior state.
    \item \emph{Resume Unconditional Reverse Diffusion}: From this modified state, we proceed with the \emph{unconditional} reverse SDE. Empirically, this yields samples that approximate the correct posterior $p(\mathbf{x}_0|\mathbf{y})$, leveraging the pretrained diffusion model's ``strong prior'' on clean images.
\end{enumerate}

The above time-continuous viewpoint leads to a one-step restoration formula.
The pretrained diffusion model typically provides a denoising network $\mathcal S_\theta(\mathbf{x}_t, t)$.
A standard DDPM-like update can be written as \cite{chung2022diffusion}:
\begin{equation}
\hat{\mathbf{x}_0}=\frac{1}{\sqrt{\bar{\alpha}(t)}}
\Bigl(
    \mathbf{x}_t
    + \bigl(1 - \bar{\alpha}(t)\bigr)\,\mathcal S_\theta(\mathbf{x}_t, t)
\Bigr)
\;\triangleq\;
\mathcal{R}_{\theta}\bigl(\mathbf{x}_t\bigr),   
\end{equation}
where $\bar{\alpha}(t)$ is a cumulative noise-related term (often denoted in diffusion literature), and $\mathcal{R}_{\theta}$ is the resulting ``restoration'' model. $\mathcal{R}_{\theta}$ and $\mathcal{S}_{\theta}$ share the same network architecture $\mathcal U$.
The output $\hat{\mathbf{x}_0}$ serves as an approximate clean image.

Since the above one-step (or short-step) restoration may still deviate from the true posterior, we apply a fine-tuning stage for it.
Specifically, we treat $\mathcal{R}_\theta$ as an initial generator and fine-tune it using GAN loss.
This allows the model to correct residual artifacts or unrealistic details, further narrowing the gap between the approximate samples and the real clean data distribution.
In conclusion, discarding the complex likelihood term and instead leveraging a pretrained diffusion prior, along with a subsequent injection of the observed data, proves to be a practical and effective approximation for image restoration.
The pretrained model's denoising capability provides an excellent initialization, while a GAN-based refinement further enhances visual fidelity, making diffusion priors inherently compelling for a wide range of restoration tasks.

\begin{figure*}[t]
    \centering
    \includegraphics[width=\linewidth]{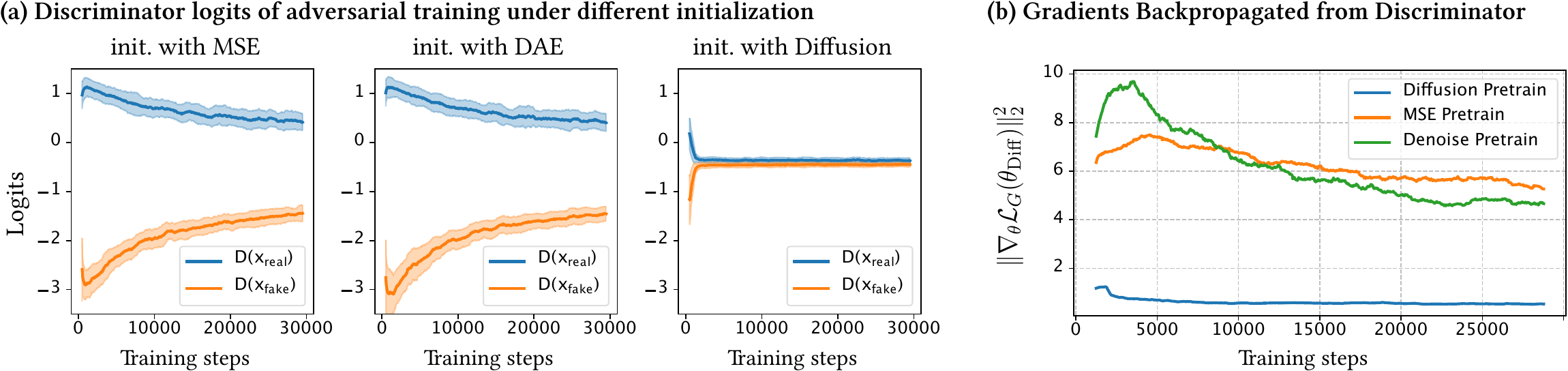}
    \caption{(a) The discriminator logits for real and generated images across training steps. Diffusion-based initialization yields rapid and stable convergence, reflecting better alignment between generated and real distributions compared to MSE and denoising autoencoder (DAE) initializations. (b) Magnitude of gradients backpropagated from the discriminator to the generator. Diffusion initialization produces consistently small gradients, highlighting improved numerical stability and efficiency during GAN post-training.}
    \Description{(a) The discriminator logits for real and generated images across training steps. Diffusion-based initialization yields rapid and stable convergence, reflecting better alignment between generated and real distributions compared to MSE and denoising autoencoder (DAE) initializations. (b) Magnitude of gradients backpropagated from the discriminator to the generator. Diffusion initialization produces consistently small gradients, highlighting improved numerical stability and efficiency during GAN post-training.}
    \label{fig:gradient}
\end{figure*}

\begin{tcolorbox}[colframe=black!50,colback=yellow!10,boxrule=0.5pt, boxsep=1pt]
(2) Why does using the diffusion model as initialization substantially improve adversarial training for restoration? 
\end{tcolorbox}

From the discussion above, we see that initializing a restoration model with diffusion weights positions it already close to the natural image space.
This intuition can be formalized through the following theorem, which quantitatively demonstrates the proximity of the diffusion-initialized distribution to the natural image space.

\medskip
\hrule
\begin{theorem}[Diffusion-to-Restoration Proximity]
\label{thm:diff2rest}
Assume the diffusion network $\mathcal U_{\theta_\mathrm{Diff}}$  whose score error on $(p_{\mathrm{data}} *k_\sigma)$ is bounded by $\varepsilon_{\mathrm{sc}}$, where $\theta_\mathrm{Diff}$ denotes the parameters of the pretrained diffusion model.
And let $k_{\mathrm{deg}}$ be the degradation kernel that satisfies \(\Delta_k := \|k_{\mathrm{deg}}-k_\sigma\|_1 \ll 1\).
Set $\mathcal R_{\theta_\mathrm{Diff}}:=\mathcal U_{\theta_\mathrm{Diff}}$, acting on $\mathbf y\sim p_y$.
Then the push-forward distribution \(p_{\theta_\mathrm{Diff}}:=\mathcal R_{\theta_\mathrm{Diff}}\,\sharp\,p_y\) obeys the quantitative bound 
\[
  W_2 \bigl(p_{\theta_\mathrm{Diff}},\,p_{\mathrm{data}}\bigr)
  \;\le\;
  C_1\,\varepsilon_{\mathrm{sc}}
  \;+\;
  C_2\,\Delta_k
  \;=: \epsilon_0,
\]
where $C_1,C_2>0$ depend only on the Lipschitz constants of
$\mathcal U_{\theta_\mathrm{Diff}}$ and the image space dimension.
\end{theorem}
\hrule
\medskip
\noindent\textit{Proof.}~
The detailed proof of Theorem~\ref{thm:diff2rest} is provided in Appendix~\ref{thm:diff2rest:proof}.
\medskip

We thus arrive at three principal corollaries based on Theorem~\ref{thm:diff2rest}:
\begin{enumerate}
  \item \textbf{Small initial gradient.}  
        The GAN gradient remains relatively small, thereby avoiding common issues such as NaNs, gradient explosion, and oscillatory updates.
  \item \textbf{Near-complete mode coverage.}  
        The generated restoration distribution already overlaps significantly with the real image distribution, substantially reducing the risk of mode collapse.
  \item \textbf{Faster convergence.}  
        Gradient descent now requires far fewer iterations to reach the target accuracy than a model trained from scratch without diffusion initialization.
\end{enumerate}
We elaborate on each of these points in detail next.

\paragraph{\textbf{Small initial gradient.}}
\label{sec:initial}
Because $p_{\theta_\mathrm{Diff}}$ is already $\epsilon_0$-close to $p_{\mathrm{data}}$ in $W_2$, the generator's first adversarial update is guaranteed to be tiny and well-behaved.
Define the optimal discriminator corresponding to a fixed generator parameter $\theta$ by \(D_\theta^\star(x)\;=\;\frac{p_{\mathrm{data}}(x)}{p_{\mathrm{data}}(x)+p_\theta(x)}\).
Let the non-saturating generator loss be
\[
\mathcal L_G(\theta)=
      -\,\mathbb E_{y\sim p_y} \bigl[\log D^\star_\theta(\mathcal R_\theta(y))\bigr].
\]
We formalize this conclusion as a lemma.

\begin{lemma}[Initial gradient bound]
\label{lem:init-grad}
Assume  
\begin{enumerate}[label=(\alph*)]
  \item The generator $\mathcal R_\theta$ is $L_g$-Lipschitz in its output argument: $\|\mathcal R_\theta(y_1)-\mathcal R_\theta(y_2)\|\le L_g\|y_1-y_2\|$.  
  \item Its parameter Jacobian at $\theta=\theta_\mathrm{Diff}$ is uniformly bounded a constant $L_J$: $\|J_{\theta_\mathrm{Diff}}(y)\|\le L_J$ for all $y$.
  \item $p_{\mathrm{data}},p_{\theta_\mathrm{Diff}}$ admit densities bounded above by a constant $M$.
\end{enumerate}
Then
\[
   \bigl\|\nabla_\theta\mathcal L_G(\theta_\mathrm{Diff})\bigr\|_2
   \;\le\;
   \sqrt2\,L_J\,\epsilon_0,
\]
where $\epsilon_0 = C_1\varepsilon_{\mathrm{sc}} + C_2\Delta_k$ as in
Theorem~\ref{thm:diff2rest}.
\end{lemma}
\noindent\textit{Proof.}~
The detailed proof of Lemma~\ref{lem:init-grad} is provided in Appendix~\ref{lem:init-grad:proof}.
\medskip

As $\nabla_\theta\mathcal L_G(\theta_\mathrm{Diff})$ is already proportional to~$\epsilon_0$, the optimization begins on a near-saddle plateau: the loss surface is almost flat for~$\theta$, while the discriminator gradient is also small for~$\phi$ (the parameters for the discriminator).
Practically, this diffusion initialization offers benefits for numerical stability.
Because the resulting Lipschitz-bounded gradients remain tiny, avoiding exploding updates or NaNs, even with aggressive learning rates.
Thus, the fine-tuning phase resembles sculpting fine details rather than a laborious search.

This conclusion is supported by our experimental results.
\figurename~\ref{fig:gradient} shows the discriminator logits (a) and the magnitude of generator gradients (b) using different initialization methods.
It can be observed that, compared to MSE initialization and denoising autoencoder (DAE) initialization, the discriminator logits for real images and generated images quickly stabilize and converge closely under diffusion-based initialization.
This behavior indicates a higher similarity between the generated distribution and the real image distribution.
Furthermore, as shown in \figurename~\ref{fig:gradient} (b), the gradient magnitudes remain consistently low under diffusion-based initialization, whereas the other initialization methods exhibit significantly higher gradient fluctuations.
This stable gradient behavior empirically validates our theoretical analysis (Lemma \ref{lem:init-grad}) that diffusion-based initialization effectively ensures numerical stability, thereby making the GAN fine-tuning process more efficient and robust.

\paragraph{\textbf{Near‐complete mode coverage.}}
\label{sec:mode}
Denote by $\{\mathcal A_k\}_{k=1}^{K}$ any measurable partition of the image space.
We show that every such region already carries almost the correct probability mass under the restoration generator initialized by a pre-trained diffusion model.

\begin{proposition}[Uniform mode‐mass bound]
\label{prop:mode}
For every $k\in\{1,\dots,K\}$ and $\epsilon_0$ defined in Theorem~\ref{thm:diff2rest}, it holds that
\[
   \bigl|
     p_{\theta_\mathrm{Diff}}(\mathcal A_k)\;-\;
     p_{\mathrm{data}}(\mathcal A_k)
   \bigr|
   \;\le\;
   \tfrac{\sqrt2}{2}\,\epsilon_0.
\]
\end{proposition}
\noindent\textit{Proof.}~
The proof of Proposition~\ref{prop:mode} is provided in Appendix~\ref{prop:mode:proof}.
\medskip

The uniform bound of Proposition~\ref{prop:mode} can be translated into the two practical advantages.
First, no semantic region can lose more than $\tfrac{\sqrt2}{2}\,\epsilon_0$ probability mass, so even the rarest modes remain represented in the generator's support.
Second, with mode collapse largely mitigated and all major modes already covered, the discriminator's gradients shift toward higher-order discrepancies.
As a result, fine-tuning focuses on refining color nuances and sharpening edges, rather than generating missing categories.

\begin{figure}[t]
    \centering
    \includegraphics[width=\linewidth]{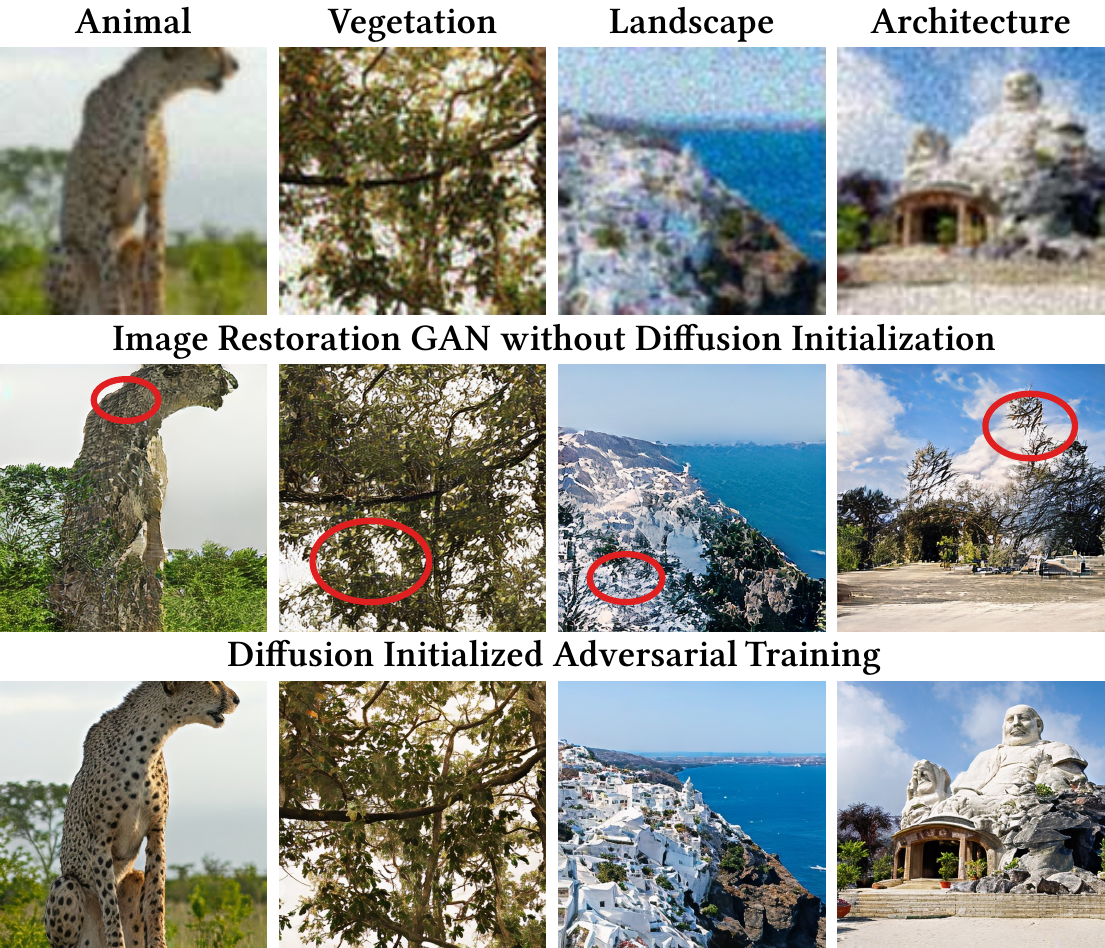}
    \caption{Visual comparison illustrating mode collapse in image restoration GANs without diffusion initialization (middle row, problematic textures highlighted) versus the improved semantic diversity achieved by the proposed diffusion-initialized adversarial training (bottom row). Please refer to the magnified view for a more detailed examination. Photo Credits: Images from the DIV2K dataset (licensed CC BY 4.0).}
    \Description{Visual comparison illustrating mode collapse in image restoration GANs without diffusion initialization (middle row, problematic textures highlighted) versus the improved semantic diversity achieved by the proposed diffusion-initialized adversarial training (bottom row). Please refer to the magnified view for a more detailed examination. Photo Credits: Images from the DIV2K dataset (licensed CC BY 4.0).}
    \label{fig:mode}
\end{figure}

A major challenge of previous GAN-based restoration methods is mode collapse during texture restoration, where models produce similar textures across semantically distinct regions.
For example, in \figurename~\ref{fig:mode}, the GAN without diffusion initialization generates visually similar textures across various semantic categories (highlighted by the red circles), including animal, vegetation, landscapes, and architectural structures.
This indicates the model's inability to distinguish different semantics, causing repetitive patterns.
In contrast, diffusion-initialized adversarial training significantly mitigates this issue, effectively covering diverse semantic modes and producing accurate, distinct textures corresponding to each image category.

\begin{figure}[t]
    \centering
    \includegraphics[width=\linewidth]{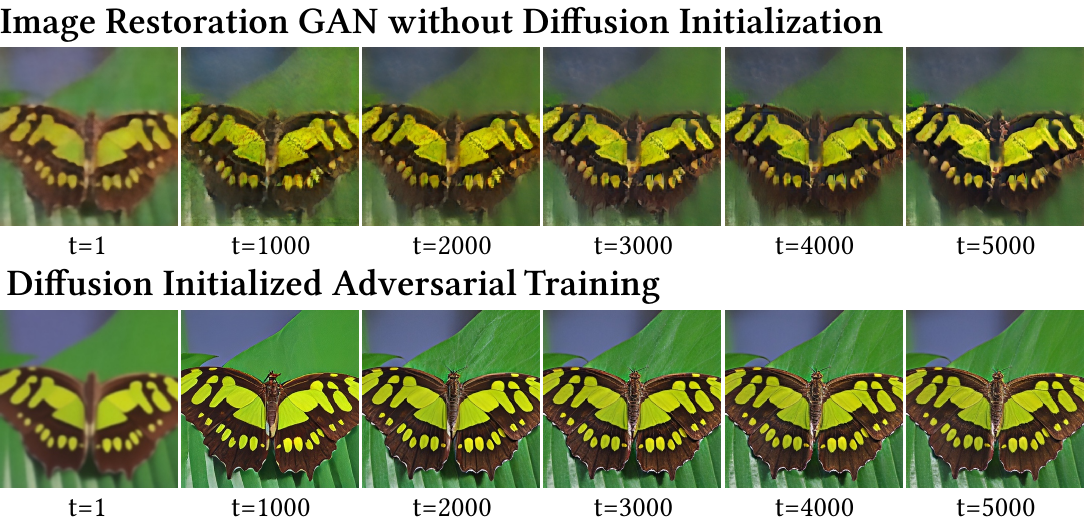}
    \caption{Comparison of restoration progress without (top) and with diffusion initialization (bottom). Diffusion initialization yields clearer, stable outputs early in training. Please refer to the magnified view for a more detailed examination. Photo Credits: Images from the DIV2K dataset (licensed CC BY 4.0).}
    \Description{Comparison of restoration progress without (top) and with diffusion initialization (bottom). Diffusion initialization yields clearer, stable outputs early in training. Please refer to the magnified view for a more detailed examination. Photo Credits: Images from the DIV2K dataset (licensed CC BY 4.0).}
    \label{fig:fast}
\end{figure}

\paragraph{\textbf{Faster convergence.}}
\label{sec:fast}
Once the initial gradient is guaranteed to be of order $\mathcal O(\epsilon_0)$, the local geometry of the generator loss implies that only a logarithmic number of gradient steps is required to close the remaining gap.

\begin{proposition}[Linear-logarithmic convergence rate]
\label{prop:linlog}
Suppose that, in a neighborhood $\mathcal N(\theta_\mathrm{Diff},r)=\{\theta:\|\theta-\theta_\mathrm{Diff}\|\le r\}$, the generator loss $\mathcal L_G$ is
\begin{itemize}
  \item \emph{$L$--smooth}: $\|\nabla \mathcal L_G(\theta_1)-\nabla \mathcal L_G(\theta_2)\|\le L\|\theta_1-\theta_2\|$,
  \item \emph{$\mu$--strongly convex}: $\mathcal L_G(\theta_2)\ge\mathcal L_G(\theta_1)
        +\langle\nabla \mathcal L_G(\theta_1),\theta_2-\theta_1\rangle
        +\tfrac{\mu}{2}\|\theta_2-\theta_1\|^2$,
\end{itemize}
with $\eta\le 1/L$ the generator step size (the standard ``safe'' choice).
Let $\theta^0=\theta_\mathrm{Diff}$ and iterate $\theta^{t+1}=\theta^t-\eta\nabla_\theta\mathcal L_G(\theta^t)$.
Denote $\delta_t:=\mathcal L_G(\theta^t)-\mathcal L_G^{\star}$, where $\mathcal L_G^{\star}$ denotes the global optimum of the generator loss under the optimal discriminator, and target accuracy $\delta_{\mathrm{tar}}>0$.
Then
\[
  t
  \;\;\ge\;\;
  ^{\ln\bigl(\tfrac{C_2\epsilon_0^{2}}{\delta_{\mathrm{tar}}}\bigr)}
       /_{\ln(1/(1-\eta\mu))}
  \;=\;
  \mathcal O\Bigl(\log\tfrac{1}{\epsilon_0}\Bigr)
\]
suffices to guarantee $\delta_t\le\delta_{\mathrm{tar}}$.
\end{proposition}
\noindent\textit{Proof.}~
The proof of Proposition~\ref{prop:linlog} is provided in Appendix~\ref{prop:linlog:proof}.
\medskip

With practical constants $L \sim 3\times10^{2}$, $\mu \sim 6\times10^{-2}$, $\eta = 1/L$, $\epsilon_0 \sim 5\times 10^{-3}$ ($\mathrm{FID}\sim 2$), $\delta_{\mathrm{tar}} = 1\times 10^{-5}$ ($\mathrm{FID}\sim 1$) and $C_2\sim 2$, Proposition \ref{prop:linlog} predicts $t\approx 8\times10^{3}$ iterations.
Although this is only a rough estimate, our experiments indicate that diffusion-initialized GAN post-training for $512\times 512$ image restoration tasks typically converges within approximately $10$k steps, consistent in order of magnitude with our theoretical predictions.
\figurename~\ref{fig:fast} illustrates the progression of image quality with the increased training steps using our method.
It can be observed that our approach already generates reasonable images at initialization point, which fits theorem \ref{thm:diff2rest}, and after just a few thousand training iterations, it produces clear and high-quality images.
In contrast, the method without initialization remains unstable and continues oscillating during the same training period.
Generally speaking, training a restoration generator from scratch with the same hyper-parameters typically demands $\ge 3\times 10^{5}$ iterations.
Both our empirical results and the accompanying theory concur that the proposed method converges far more rapidly.

\section{Other Implementation Details}
The previous discussion focused on the core idea behind our approach.
Next, we detail how to practically build a large-scale image restoration model based it.

\subsection{Diffusion Model Selection}
Within our framework, the diffusion model fundamentally determines the baseline performance.
Naturally, employing larger and more advanced diffusion models leads to superior results \cite{yu2024scaling}.
In our experiments, we investigate four diffusion models: SD2 \cite{rombach2022high} (0.8B parameters), SDXL \cite{podell2023sdxl} (2.6B parameters), SD3 \cite{esser2024scaling} (8B parameters), and Flux \cite{flux2024dev} (12B parameters).
Leveraging large-scale models for image restoration has always been a significant goal \cite{yu2024scaling,yu2025unicon}.
As our method does not require additional control adapters, we are uniquely positioned to efficiently exploit these large-scale diffusion models, achieving significantly reduced computational requirements compared to previous methods.
In contrast, previous approaches have largely struggled to leverage large-scale diffusion models such as SD3 or Flux, due to prohibitive computational costs.

It is also important to note that the training methods for advanced diffusion models have evolved beyond the original NCSN diffusion model \cite{song2019generative}.
This discrepancy arises because the training of modern diffusion models involves various engineering constraints related to datasets, optimization techniques, and other practical considerations.
On the other hand, diffusion models often exhibit robustness and impressive capabilities in practice, frequently surpassing theoretical predictions.
Various diffusion variants, such as EDM \cite{karras2022elucidating} and ReFlow \cite{lee2024improving}, have introduced modifications that relax theoretical constraints, achieving remarkable practical improvements.
Employing these advanced, larger-scale diffusion models further enhances the robustness and effectiveness of our image restoration models.
We can also benefit from the additional control capabilities of these models.

\begin{figure}
    \centering
    \includegraphics[width=\linewidth]{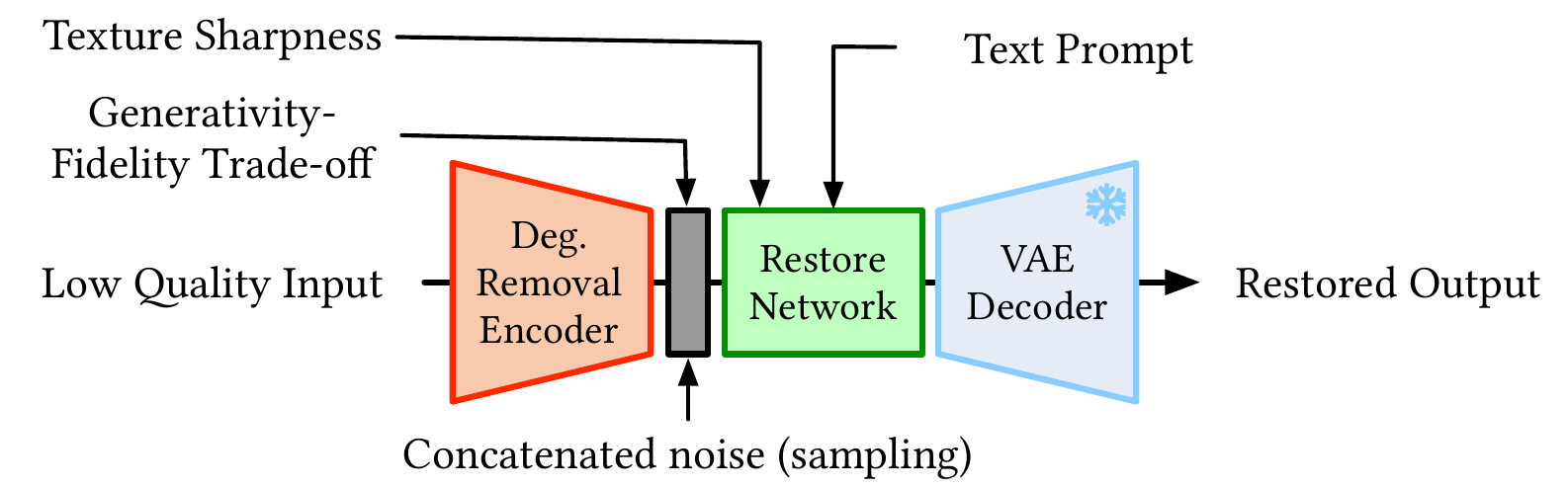}
    \caption{Illustration of controllability. Our method supports flexible manipulation through textual prompts, texture richness adjustment, generativity-fidelity trade-offs, and random sampling capability.}
    \label{fig:control-fig}
\end{figure}

\begin{figure*}
    \centering
    \includegraphics[width=\linewidth]{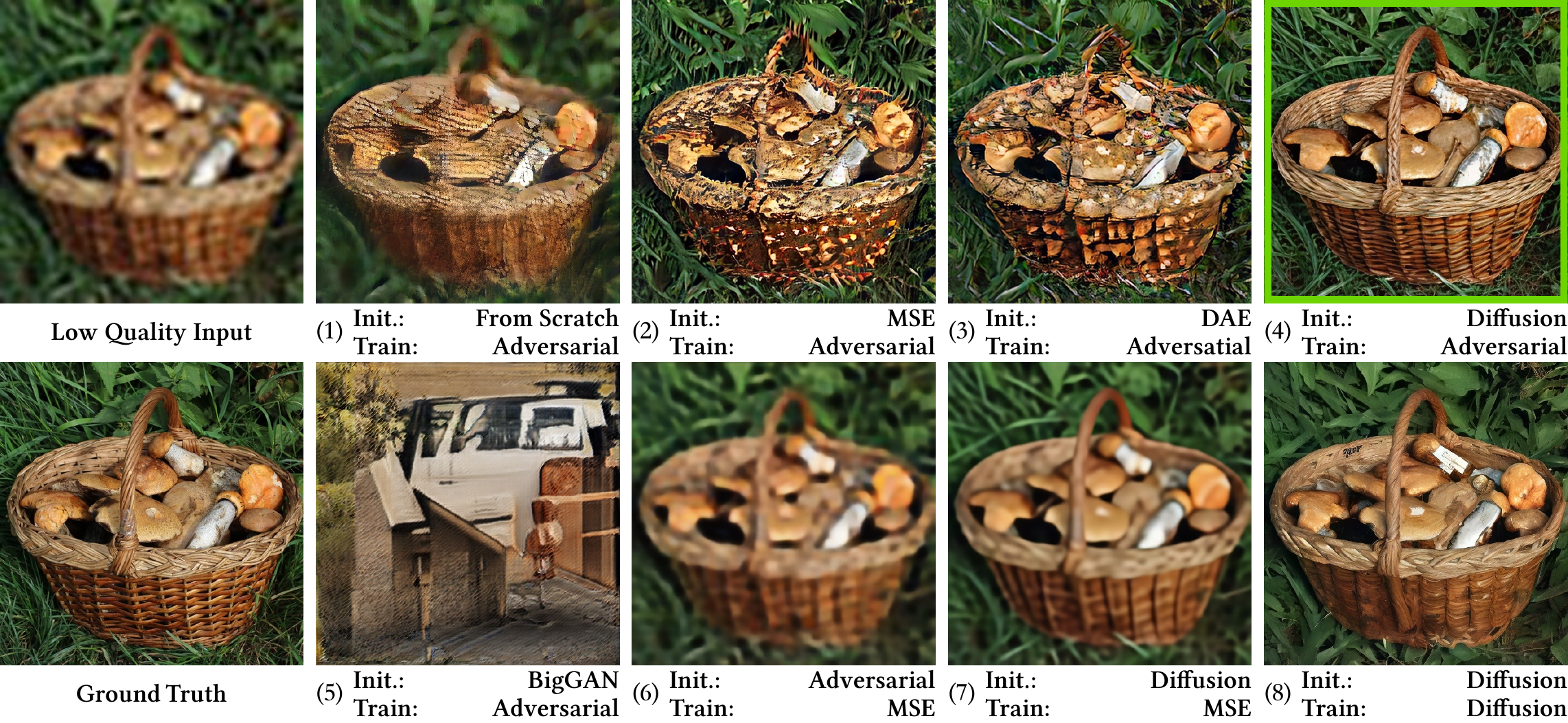}
    \caption{Comparison of various initialization and pretraining strategies for adversarial trained image restoration. Diffusion-based initialization (our method) notably enhances image quality and training stability compared to direct GAN training, MSE pretraining, and denoising autoencoder initialization. Photo Credits: Images from the DIV2K dataset (licensed CC BY 4.0).}
    \Description{Comparison of various initialization and pretraining strategies for adversarial trained image restoration. Diffusion-based initialization (our method) notably enhances image quality and training stability compared to direct GAN training, MSE pretraining, and denoising autoencoder initialization. Photo Credits: Images from the DIV2K dataset (licensed CC BY 4.0).}
    \label{fig:discussion}
\end{figure*}

\begin{figure*}
    \centering
    \includegraphics[width=\linewidth]{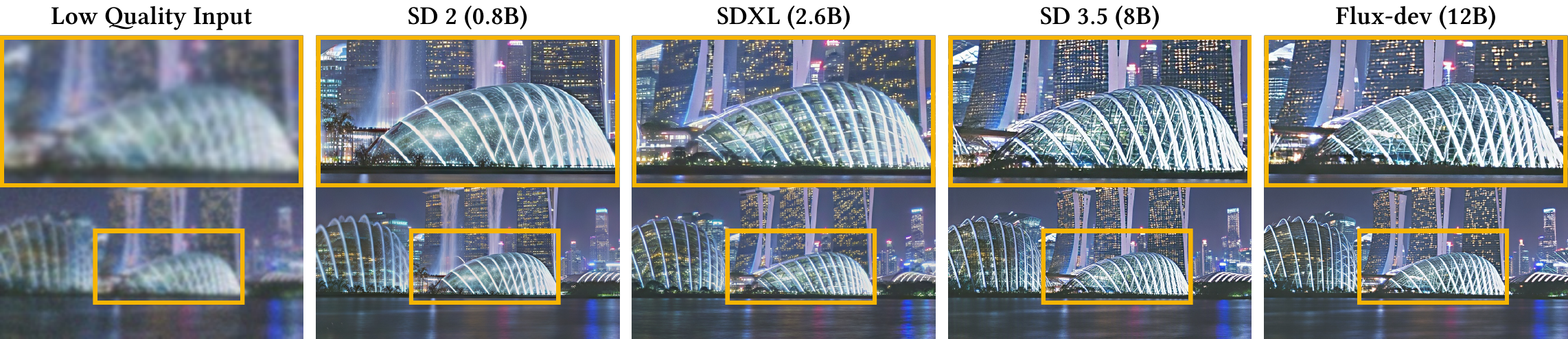}
    \caption{Performance comparison when initializing with diffusion models of varying sizes and capabilities. Larger and more advanced pretrained diffusion models enhance restoration quality, demonstrating better accuracy in score function approximation and improved handling of complex image structures. Photo Credits: Images from the DIV2K dataset (licensed CC BY 4.0).}
    \Description{Performance comparison when initializing with diffusion models of varying sizes and capabilities. Larger and more advanced pretrained diffusion models enhance restoration quality, demonstrating better accuracy in score function approximation and improved handling of complex image structures. Photo Credits: Images from the DIV2K dataset (licensed CC BY 4.0).}
    \label{fig:base-model}
\end{figure*}

\begin{figure*}
    \centering
    \includegraphics[width=\linewidth]{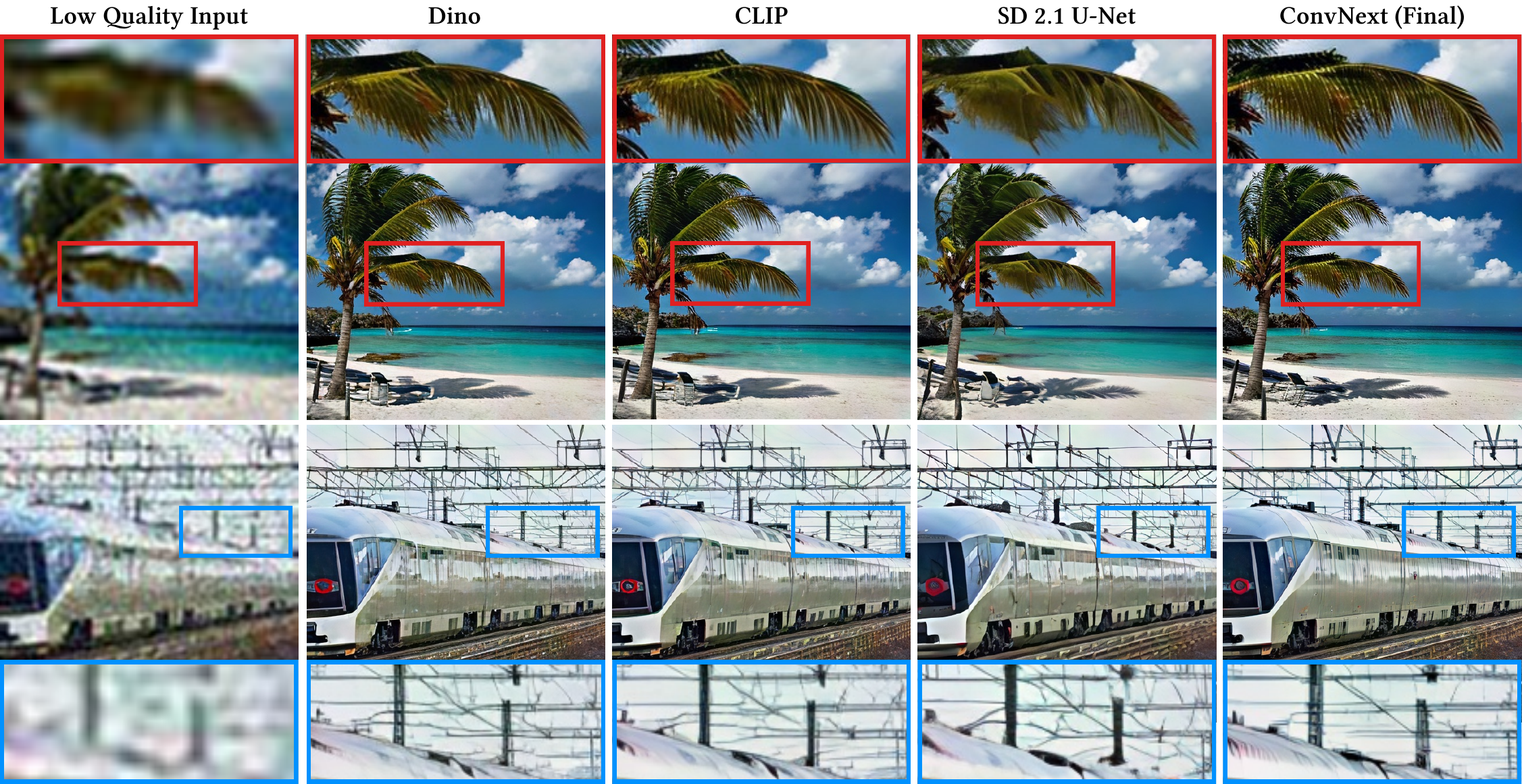}
    \caption{Comparison of restoration models guided by discriminators using different pretrained vision backbones (DINO, CLIP, diffusion U-Net, and ConvNeXt). Photo Credits: Images from the DIV2K dataset (licensed CC BY 4.0).}
    \Description{Comparison of restoration models guided by discriminators using different pretrained vision backbones (DINO, CLIP, diffusion U-Net, and ConvNeXt). Photo Credits: Images from the DIV2K dataset (licensed CC BY 4.0).}
    \label{fig:D}
\end{figure*}

\subsection{Conditional Image Restoration}
\label{sec:control}
Diffusion models impact the field with their ability to generate images conditioned on textual prompts \cite{ramesh2021zero,ramesh2022hierarchical}.
Leveraging this capability, recent diffusion-based image restoration methods have also incorporated textual control \cite{yu2024scaling}.
Beyond textual prompts alone, controllability also encompasses managing the richness of generated textures, balancing between generation and fidelity to input images, and introducing randomness for sampling diverse outputs.
These functionalities empower users to dynamically adjust restoration results according to image content and personal preferences.
Our framework accommodates these forms of control.
An illustration can be found in \figurename~\ref{fig:control-fig}.

\paragraph{Textual Prompts}
The diffusion model used in the proposed method inherently supports textual inputs.
We find that this capability remains effective within our diffusion-initialized adversarial training pipeline.
Specifically, during training, we gather textual annotations for all training images and provide them during training.
The approach for obtaining these textual descriptions follows the methodology of \citet{yu2024scaling}.
Specifically, we utilize the LLaVA model \cite{liu2023visual} to generate textual annotations for each image in the training set.
At test time, we again employ the same model to automatically predict textual prompts for the test images.
These generated captions then serve as prompts during image restoration, fully automating the prompt-based control mechanism without manual intervention.
Text prompts let image restorers (i) recover missing details by embedding a semantic description of the degraded input, (ii) fill unrecoverable regions according to user-specified content, and (iii) adjust fidelity and style through natural-language guidance.

\paragraph{Texture Richness}
Generative models typically produce realistic textures, yet different images vary a lot in texture density.
Imposing a fixed texture richness across all images can lead to two critical issues: over-amplifying textures in images with sparse textures and under-representing textures in those with abundant details.
To mitigate these drawbacks, we introduce a control metric, termed \emph{Texture Richness}, derived from statistical properties of the image's Laplacian value.
Richer textures exhibit pronounced and rapidly varying high-frequency details, resulting in greater amplitude and fluctuation in the Laplacian response.
During training, we explicitly provide the ground truth texture statistics to guide the model in producing appropriate texture richness for each image.
At inference time, users can dynamically adjust this metric, effectively controlling the global texture density of restored images.

\paragraph{Generativity-Fidelity Trade-off}
Another control dimension for practical restoration is balancing image fidelity against generative flexibility.
Severely degraded images may contain irrecoverable details, thus strictly adhering to the input can introduce undesirable artifacts.
Allowing models with greater generative flexibility in such scenarios often yields improved visual quality, as it prevents models from matching ambiguous or lost details.
We introduce noise into the encoded representation obtained from the degradation pre-removal encoder, partially obscuring the original input signal.
By adjusting this noise injection at test time, users can modulate the desired balance between strict restoration and generative freedom.

\paragraph{Random Sampling}
The restoration output from generative models may not always align perfectly with user expectations, as multiple plausible restorations can correspond to a single degraded input.
Providing multiple diverse restoration results through random sampling adds practical value.
Since we introduce noise to the encoded representation when achieving the generativity-fidelity tradeoff, this inherently adds randomness to our reconstruction results.
By sampling different noises, we can achieve a wide variety of reconstruction effects.
Moreover, as the noise scale increases, the sampling space becomes larger.

\begin{figure*}
    \centering
    \includegraphics[width=\linewidth]{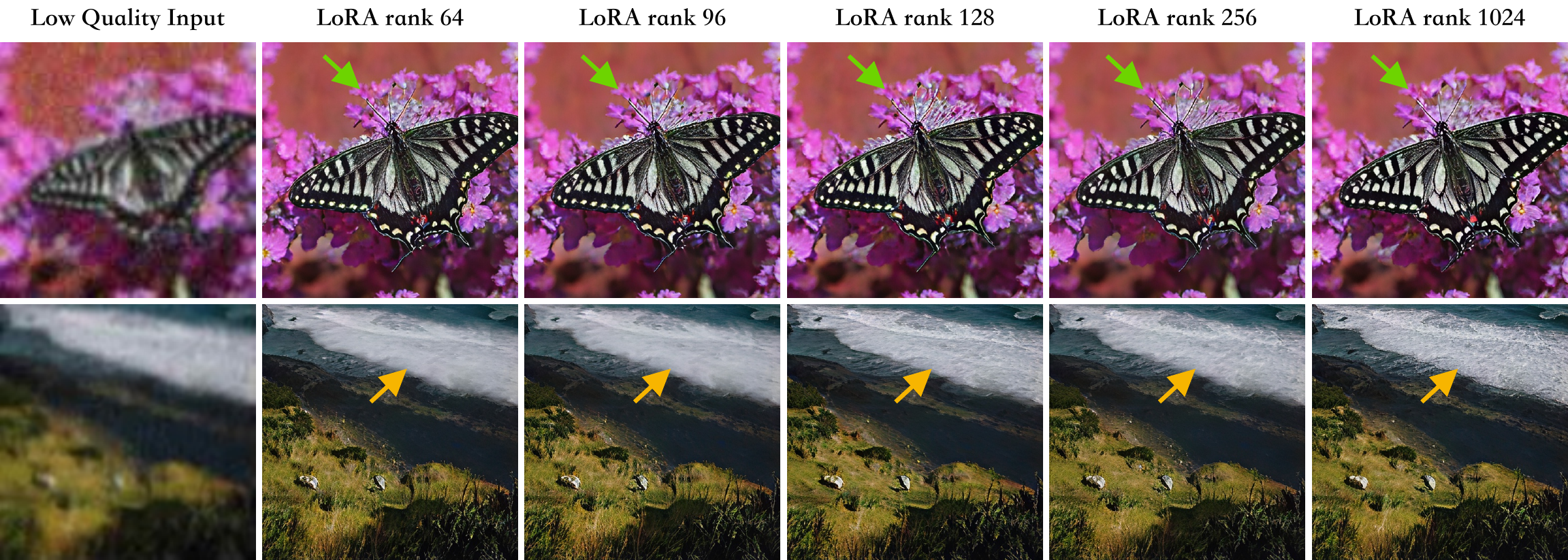}
    \caption{Visual comparison of image restoration results under different LoRA ranks. Increasing the LoRA rank expands model capacity. However, selecting an excessively high rank may yield diminishing returns relative to the computational cost. Photo Credits: Images from the DIV2K dataset (licensed CC BY 4.0).}
    \Description{Visual comparison of image restoration results under different LoRA ranks. Increasing the LoRA rank expands model capacity. However, selecting an excessively high rank may yield diminishing returns relative to the computational cost. Photo Credits: Images from the DIV2K dataset (licensed CC BY 4.0).}
    \label{fig:rank}
\end{figure*}

\begin{figure}
    \centering
    \includegraphics[width=\linewidth]{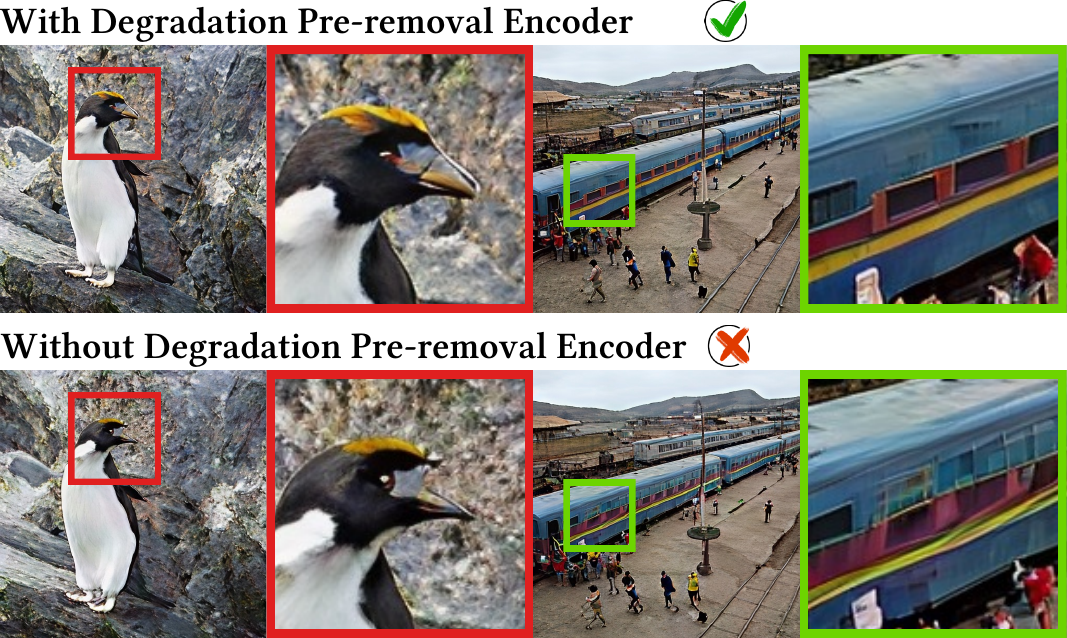}
    \caption{Comparison of restoration results with and without encoder-based degradation pre-removal. Without degradation pre-removal, the encoder may misinterpret degraded image content, introducing noticeable artifacts and compromising restoration quality. Incorporating degradation pre-removal strategy effectively mitigates these issues. Photo Credits: Images from the DIV2K dataset (licensed CC BY 4.0).}
    \Description{Comparison of restoration results with and without encoder-based degradation pre-removal. Without degradation pre-removal, the encoder may misinterpret degraded image content, introducing noticeable artifacts and compromising restoration quality. Incorporating degradation pre-removal strategy effectively mitigates these issues. Photo Credits: Images from the DIV2K dataset (licensed CC BY 4.0).}
    \label{fig:pre-denoise}
\end{figure}

\subsection{Model Training}
The overall training dataset comprises approximately 20 million high-quality image patches with associated textual descriptions and an additional 70 thousand face images.
To synthesize degraded inputs, we adopt the same degradation pipeline from Real-ESRGAN \cite{wang2021real,yu2024scaling}. 
When initialized with the SD2 model \cite{rombach2022high}, we resize input images to a resolution of $512 \times 512$.
We initialize the discriminator with a pretrained ConvNeXt \cite{liu2022convnet} model to leverage its capability in extracting detailed image features.
For training, we set the LoRA rank to 64, with the loss weights configured as follows: adversarial loss, LPIPS \cite{zhang2018unreasonable} perceptual loss, and MSE loss with weights 0.5, 5, and 1, respectively.
Training is conducted using the AdamW optimizer \cite{kingma2014adam,loshchilov2017decoupled}.
Initially, we use a batch size of 384 and set the learning rate of both the generator and the discriminator to $1\times10^{-5}$ for 10k steps.
Subsequently, we apply gradient accumulation to effectively increase the batch size to 1536, reducing the learning rate of both the generator and discriminator to $5\times10^{-6}$ for an additional 10k steps.
The entire training process is executed on 64 Nvidia A6000 GPUs.
Exponential moving average (EMA) \cite{polyak1992acceleration} weights are maintained during training with an EMA decay factor of 0.999.
For initialization with SDXL, we crop images to $512\times512$ pixels to reduce GPU memory usage.
For Flux and SD3, input images are resized to a resolution of $1024 \times 1024$.
During testing, we utilize the EMA weights.
Input images larger than the trained resolution are divided into patches matching the training resolution, processed individually, and then stitched together to reconstruct the final restored output.

\begin{figure*}[t]
    \centering
    \begin{minipage}[t]{0.54\linewidth}
        \includegraphics[width=1.\linewidth]{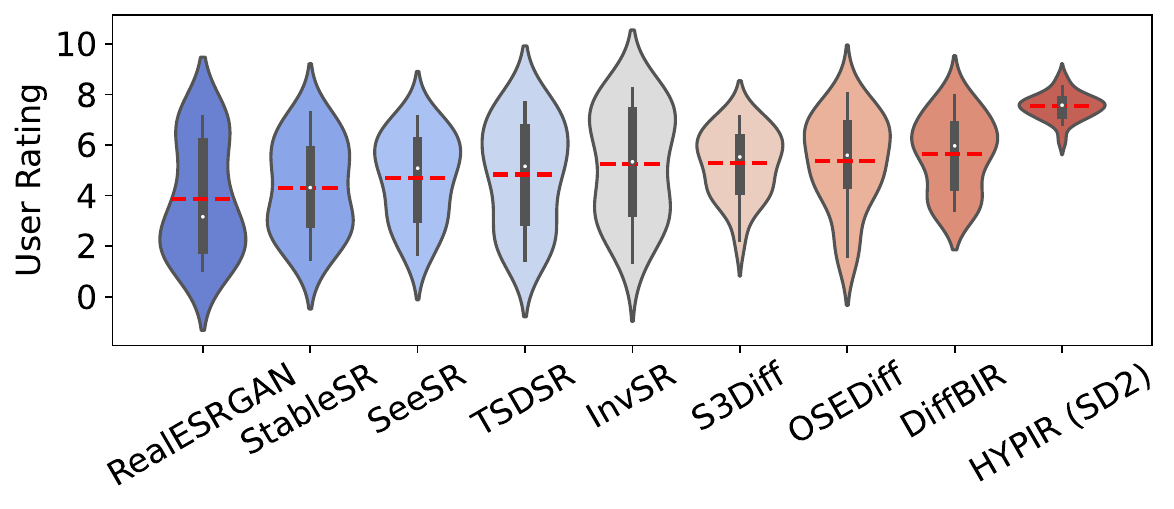}
        \captionof{figure}{User study results (Round I) comparing lightweight restoration models derived from the SD2 diffusion architecture. Our proposed method, HYPIR (SD2), achieves the highest perceptual ratings.}
        \Description{User study results (Round I) comparing lightweight restoration models derived from the SD2 diffusion architecture. Our proposed method, HYPIR (SD2), achieves the highest perceptual ratings.}
        \label{fig:user_round1}
    \end{minipage}
    \hfill
    \begin{minipage}[t]{0.42\linewidth}
        \includegraphics[width=1.\linewidth]{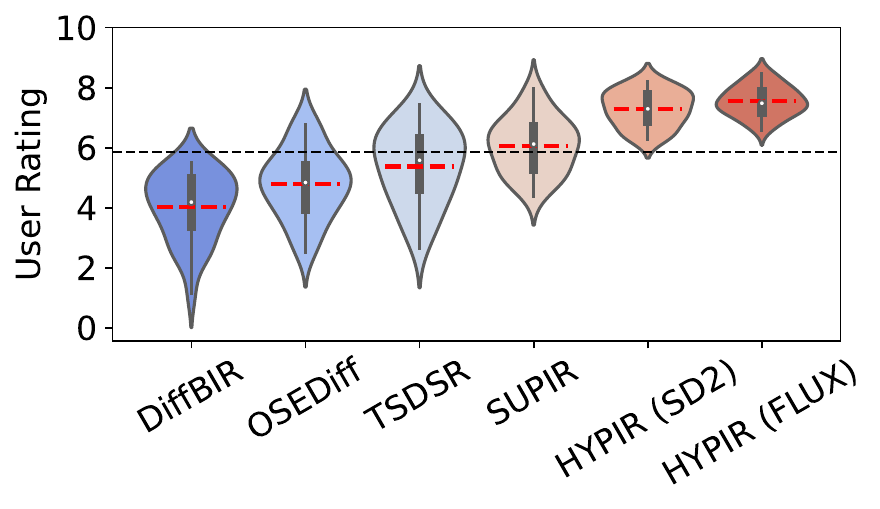}
        \captionof{figure}{User study results (Round II), extending the comparison to larger-scale models including SUPIR (SDXL) and our HYPIR variant based on Flux.}
        \Description{User study results (Round II), extending the comparison to larger-scale models including SUPIR (SDXL) and our HYPIR variant based on Flux.}
        \label{fig:user_round2}
    \end{minipage}
\end{figure*}

\begin{figure}
    \centering
    \includegraphics[width=\linewidth]{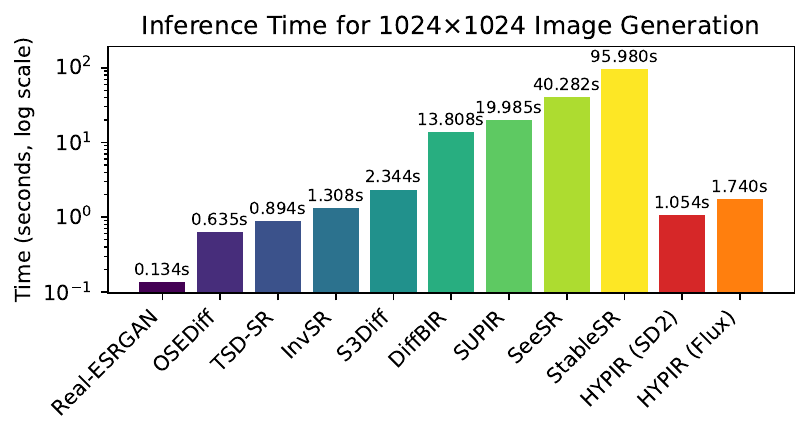}
    \caption{Comparison of inference times for generating $1024\times1024$ restored images across various methods (averaged over 100 test images). The vertical axis is shown on a logarithmic scale. Our approach achieves inference speeds comparable to other single-step models, despite employing significantly larger models (e.g., Flux, 12B parameters), while consistently delivering superior restoration quality.}
    \Description{Comparison of inference times for generating $1024\times1024$ restored images across various methods (averaged over 100 test images). The vertical axis is shown on a logarithmic scale. Our approach achieves inference speeds comparable to other single-step models, despite employing significantly larger models (e.g., Flux, 12B parameters), while consistently delivering superior restoration quality.}
    \label{fig:time}
\end{figure}

\section{Experiments}

\subsection{Ablation Studies}
We verify a series of key design choices and influential factors that critically contribute to the success of our proposed method.
All ablation studies adopt SD2 \cite{rombach2022high} as the initial diffusion model.

\subsubsection{Compare to Other Initialization and Training Methods}
GAN-based methods for image restoration have been extensively studied for many years.
It is now widely acknowledged that effective training of restoration GANs generally requires a good initialization, often achieved by pre-training the network on a pixel-wise loss (e.g., MSE loss) for the corresponding task \cite{wang2018esrgan}.
On the other hand, diffusion models are not the only models considered to learn a data distribution's score function; denoising autoencoders (DAEs) have also been theoretically identified as learning the gradient of the data distribution \cite{vincent2011connection}.
Consequently, denoising autoencoders can similarly serve as initialization methods for restoration GAN models.
We empirically investigate various initialization approaches to highlight their differences.

\figurename~\ref{fig:discussion} compares the outcomes of different initialization strategies and training strategies.
All methods, except \figurename~\ref{fig:discussion} (5), use the same network architecture.
Several observations can be made.
\figurename~\ref{fig:discussion} (1) indicates directly training a restoration GAN from scratch suffers from mode collapse and training instability.
\figurename~\ref{fig:discussion} (2) indicates initializing the network with MSE and DAE pretraining clearly improves the visual quality but does not fully resolve the inherent difficulties in GAN training.
In contrast, the diffusion-based initialization approach generates images of significantly higher quality, as shown in \figurename~\ref{fig:discussion} (4).
Notably, the image quality achieved by our approach is comparable to that of the proven and effective Diffusion Pretraining + Diffusion Fine-tuning pipeline illustrated in \figurename~\ref{fig:discussion} (8) (similar to DiffBIR and SUPIR), yet our method avoids the computationally expensive iterative generation required by these methods.
Additionally, we demonstrate that directly initializing a GAN-based generative model, such as BigGAN \cite{brock2018large}, for image restoration tasks poses considerable challenges, as illustrated in \figurename~\ref{fig:discussion} (5).
Moreover, even after effective initialization, employing only traditional image restoration objectives (MSE loss) for fine-tuning (cases of \figurename~\ref{fig:discussion} (6) and (7)) fails to fully leverage the learned generative priors.
This comparison demonstrates the effectiveness and practical advantages of our proposed diffusion-based initialization method for image restoration.

\subsubsection{Discriminator Design}
The discriminator plays a crucial role in training stability and ultimately determines the upper bound of model performance.
Traditionally, discriminators are trained from scratch, which often leads to instability during training.
Recent findings indicate that utilizing discriminators based on pretrained backbones significantly enhances training stability \cite{kumari2022ensembling}.
Inspired by this, we adopt a similar strategy and investigate four representative backbones: DINO \cite{caron2021emerging}, CLIP \cite{radford2021learning}, diffusion-based U-Net \cite{dhariwal2021diffusion}, and ConvNeXt \cite{liu2022convnet}.
DINO employs self-supervised learning via self-distillation, aiming to produce robust and discriminative image representations \cite{caron2021emerging}.
CLIP, trained on large-scale image-text datasets through contrastive learning, excels at capturing rich semantic information \cite{radford2021learning}.
The diffusion-based U-Net has been demonstrated to be a robust feature extractor due to its denoising-based generative training paradigm \cite{dhariwal2021diffusion}.
Finally, ConvNeXt, a modern convolutional backbone, efficiently extracts fine-grained features and naturally accommodates inputs of arbitrary sizes, making it particularly suitable for high-resolution training \cite{liu2022convnet}.

\figurename~\ref{fig:D} presents a comparison of these discriminators.
Both DINO and CLIP require resizing images to a fixed resolution ($224\times224$) for feature extraction, limiting their ability to supervise high-frequency details and introducing some noisy patterns.
The U-Net in the diffusion model is primarily a generative model and is not very suitable for direct discriminator design.
In contrast, ConvNeXt processes images at their restoration outcome's original resolution without resizing, significantly enhancing its ability to preserve detailed textures.
Consequently, restorations guided by ConvNeXt exhibit richer and more precise image textures.

\begin{table*}[t]
  \footnotesize
  \centering
  \caption{Quantitative results on DIV2K dataset (left) and RealPhoto60 dataset (right). The best and second best results are \textbf{bold} and \underline{underlined}, respectively.}
  \label{tab:res_quantitative}
  \begin{minipage}[t]{0.58\textwidth}
    \setlength\tabcolsep{3.9pt}
    \centering
    \vspace{0.2em}
    \begin{tabular}{l|ccc|ccccc}
      \toprule
      \textbf{Model} & \textbf{PSNR} & \textbf{SSIM} & \textbf{LPIPS}$\downarrow$ & \textbf{NIQE}$\downarrow$ & \textbf{MUSIQ} & \textbf{MANIQA} & \textbf{CLIP-IQA} & \textbf{DeQA} \\
      \midrule
      RealESRGAN   & 23.52 & \textbf{0.6298} & 0.2784 & 4.102 & 69.40 & 0.4832 & 0.6744 & 3.678 \\
      StableSR     & \underline{23.64} & 0.6071 & 0.2761 & 3.958 & 64.99 & 0.3902 & 0.5888 & 3.354 \\
      DiffBIR      & \textbf{23.80} & 0.5988 & 0.2726 & 3.904 & 69.84 & 0.4705 & 0.6398 & 3.770 \\
      InvSR        & 23.28 & 0.6123 & 0.2852 & 4.200 & 68.91 & 0.4828 & 0.6616 & 3.725 \\
      SeeSR        & 23.41 & 0.6090 & 0.2295 & 3.855 & 72.73 & 0.5133 & 0.6877 & 4.003 \\
      TSDSR        & 21.89 & 0.5679 & 0.2358 & \textbf{3.730} & \textbf{73.36} & 0.5791 & \underline{0.7470} & 3.872 \\
      OSEDiff      & 23.32 & \underline{0.6160} & \underline{0.2260} & 4.026 & 71.77 & 0.5050 & 0.6934 & \underline{4.023} \\
      \midrule
      Ours (SD2) & 22.16 & 0.5877 & 0.2318 & \underline{3.832} & 72.58 & \underline{0.5838} & 0.7467 & \textbf{4.056} \\
      Ours (Flux)  & 21.65 & 0.5982 & \textbf{0.2022} & 4.004 & \underline{73.07} & \textbf{0.5966} & \textbf{0.7555} & 3.959 \\
      \bottomrule
    \end{tabular}
  \end{minipage}
  \hfill
  \begin{minipage}[t]{0.41\textwidth}
    \setlength\tabcolsep{4pt}
    \centering
    \vspace{0.2em}
    \begin{tabular}{l|ccccc}
      \toprule
      \textbf{Model} & \textbf{NIQE}$\downarrow$ & \textbf{MUSIQ} & \textbf{MANIQA} & \textbf{CLIP-IQA} & \textbf{DeQA} \\
      \midrule
      RealESRGAN   & 5.294 & 60.38 & 0.4527 & 0.5812 & 3.313 \\
      StableSR     & 4.176 & 61.88 & 0.3679 & 0.5995 & 3.355 \\
      DiffBIR      & 4.096 & 64.67 & 0.4506 & 0.6327 & 3.615 \\
      InvSR        & 4.367 & 67.38 & 0.4846 & 0.6916 & 3.765 \\
      SeeSR        & \textbf{3.579} & 72.52 & 0.5120 & 0.7125 & 4.042 \\
      TSDSR        & 3.646 & \textbf{74.02} & \underline{0.5937} & \textbf{0.8148} & \underline{4.104} \\
      OSEDiff      & 3.868 & 71.44 & 0.5217 & 0.7339 & \textbf{4.171} \\
      \midrule
      Ours (SD2) & 3.713 & 72.50 & 0.5834 & \underline{0.7886} & \textbf{4.171} \\
      Ours (Flux)  & \underline{3.640} & \underline{73.12} & \textbf{0.5947} & 0.7871 & 3.965 \\
      \bottomrule
    \end{tabular}
  \end{minipage}
\end{table*}

\subsubsection{LoRA Training Ranks}
Given that many contemporary diffusion models are extremely large-scale, performing full-parameter training on such large networks is expensive.
On the other hand, as previously demonstrated, our approach only needs to update a small subset of parameters.
The used LoRA \cite{hu2022lora} strategy reduces the number of trainable parameters, improving accessibility and practicality.
We further investigate how different LoRA ranks influence the restoration quality.
\figurename~\ref{fig:rank} illustrates these results, showing that increasing the LoRA rank and thus expanding the model's trainable capacity generally enhances performance to some extent.
This observation suggests that the subsequent adversarial training substantially contributes to improving visual quality.

\subsubsection{Degradation Pre-removal}
We propose fine-tuning the encoder to explicitly address image degradation, enabling it to mitigate such degradations directly during the encoding phase.
Our experiments demonstrate the effectiveness of this design.
\figurename~\ref{fig:pre-denoise} compares restoration results obtained with and without this degradation pre-removal strategy.
Although it is possible to produce restored images without degradation pre-removal, the encoder tends to misinterpret certain degraded image contents, leading to noticeable artifacts and reduced restoration quality.
By contrast, incorporating degradation pre-removal improves the clarity and accuracy of the restoration results.

\begin{figure*}
    \centering
    \includegraphics[width=\linewidth]{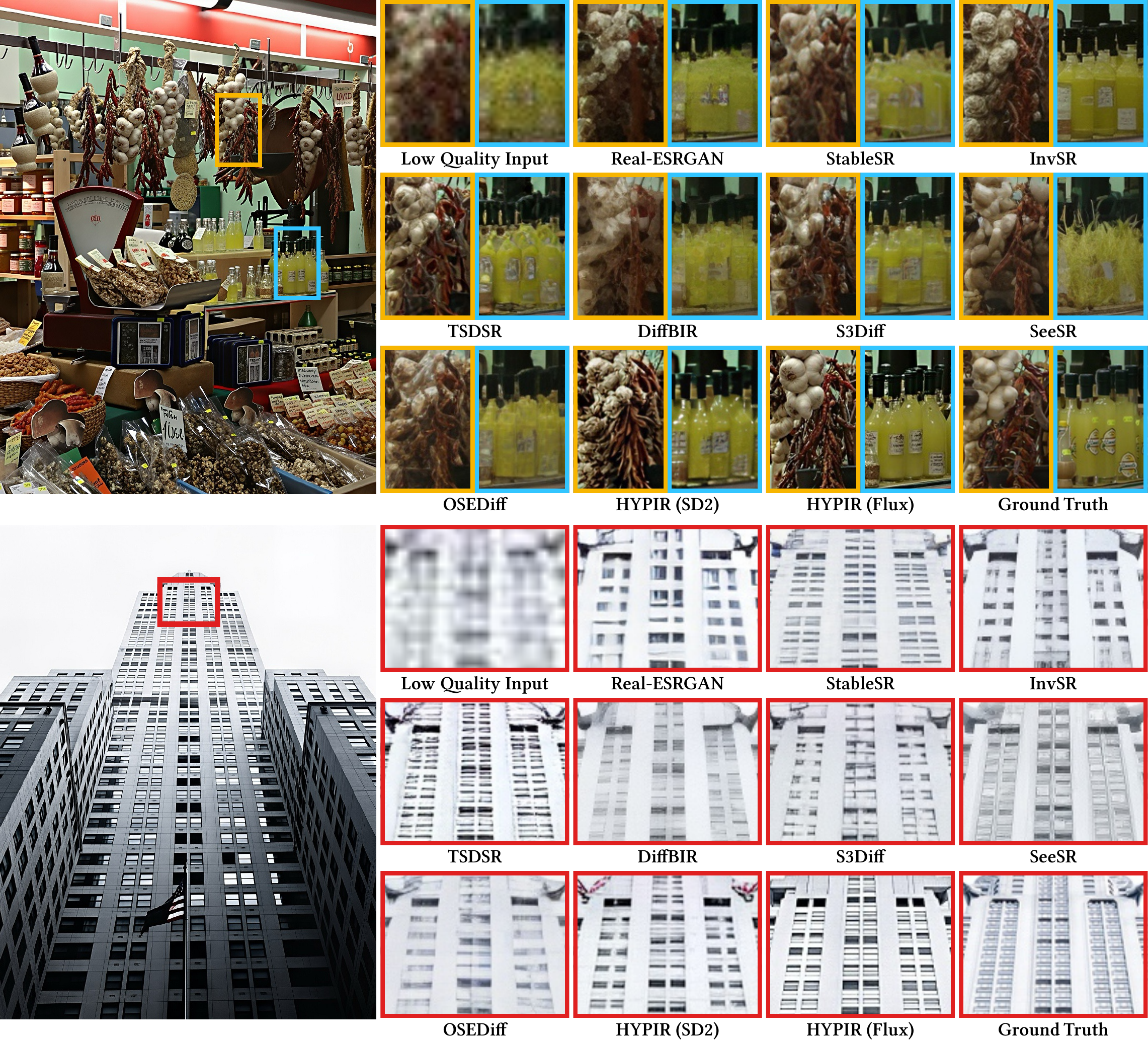}
    \caption{Qualitative comparison on synthetic test data. The SD2-based HYPIR generates superior textures compared to existing methods, while Flux-based HYPIR achieves even clearer structural restoration. Note the finely restored details of garlic and glass bottles in the first example, which other methods fail to recover clearly. Photo Credits: Images from the DIV2K dataset (licensed CC BY 4.0).}
    \Description{Qualitative comparison on synthetic test data. The SD2-based HYPIR generates superior textures compared to existing methods, while Flux-based HYPIR achieves even clearer structural restoration. Note the finely restored details of garlic and glass bottles in the first example, which other methods fail to recover clearly. Photo Credits: Images from the DIV2K dataset (licensed CC BY 4.0).}
    \label{fig:comparison}
\end{figure*}

\begin{figure*}
    \centering
    \includegraphics[width=\linewidth]{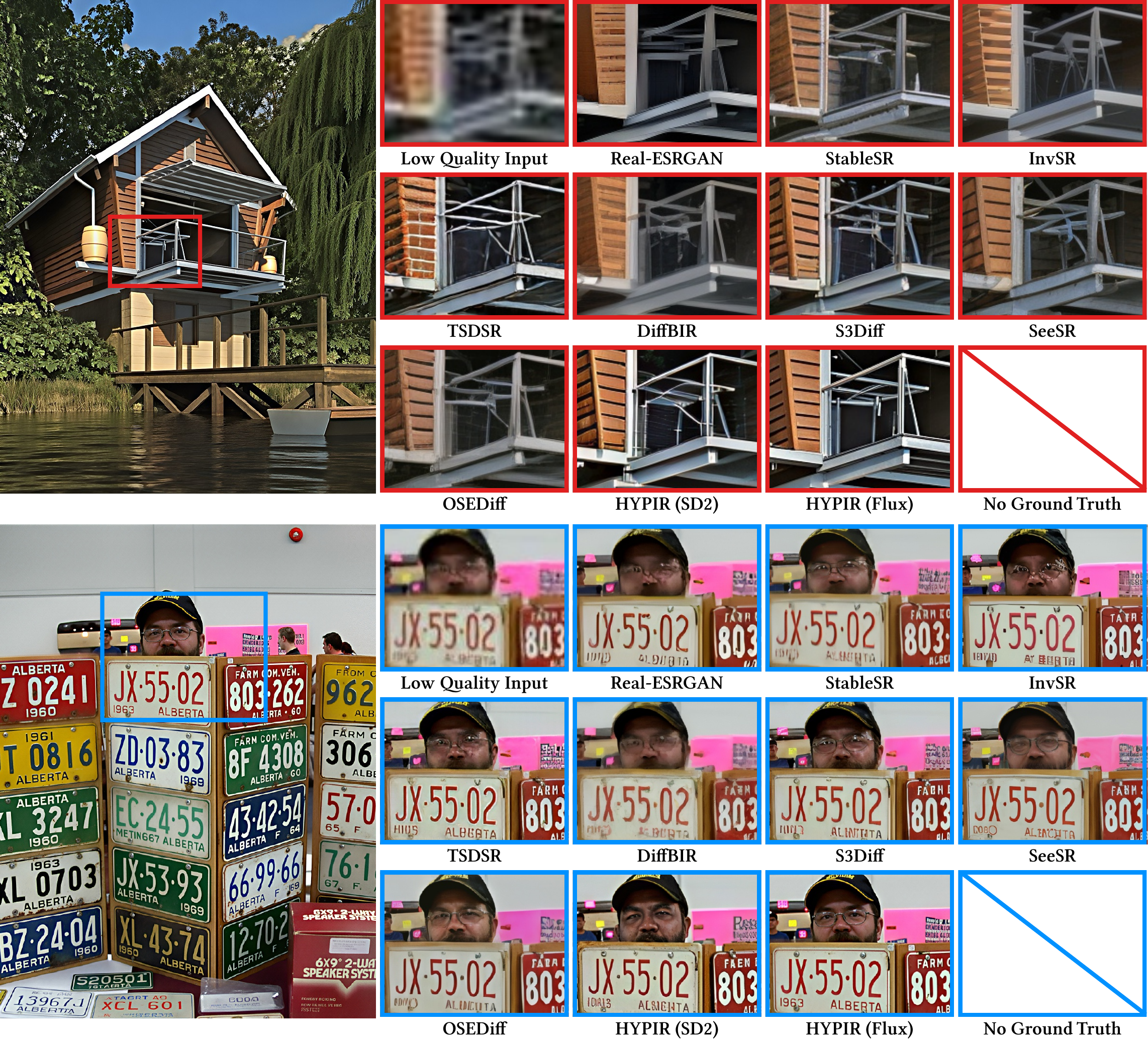}
    \caption{Restoration results on diverse real-world images, highlighting HYPIR's excellent ability to recover detailed facial features, clear textual content, and accurate architectural structures. Photo Credits: Images from the RealPhoto60 and RealLR200 datasets (licensed CC BY 4.0).}
    \Description{Restoration results on diverse real-world images, highlighting HYPIR's excellent ability to recover detailed facial features, clear textual content, and accurate architectural structures. Photo Credits: Images from the RealPhoto60 and RealLR200 datasets (licensed CC BY 4.0).}
    \label{fig:real-comparison}
\end{figure*}

\begin{figure*}
    \centering
    \includegraphics[width=0.9\linewidth]{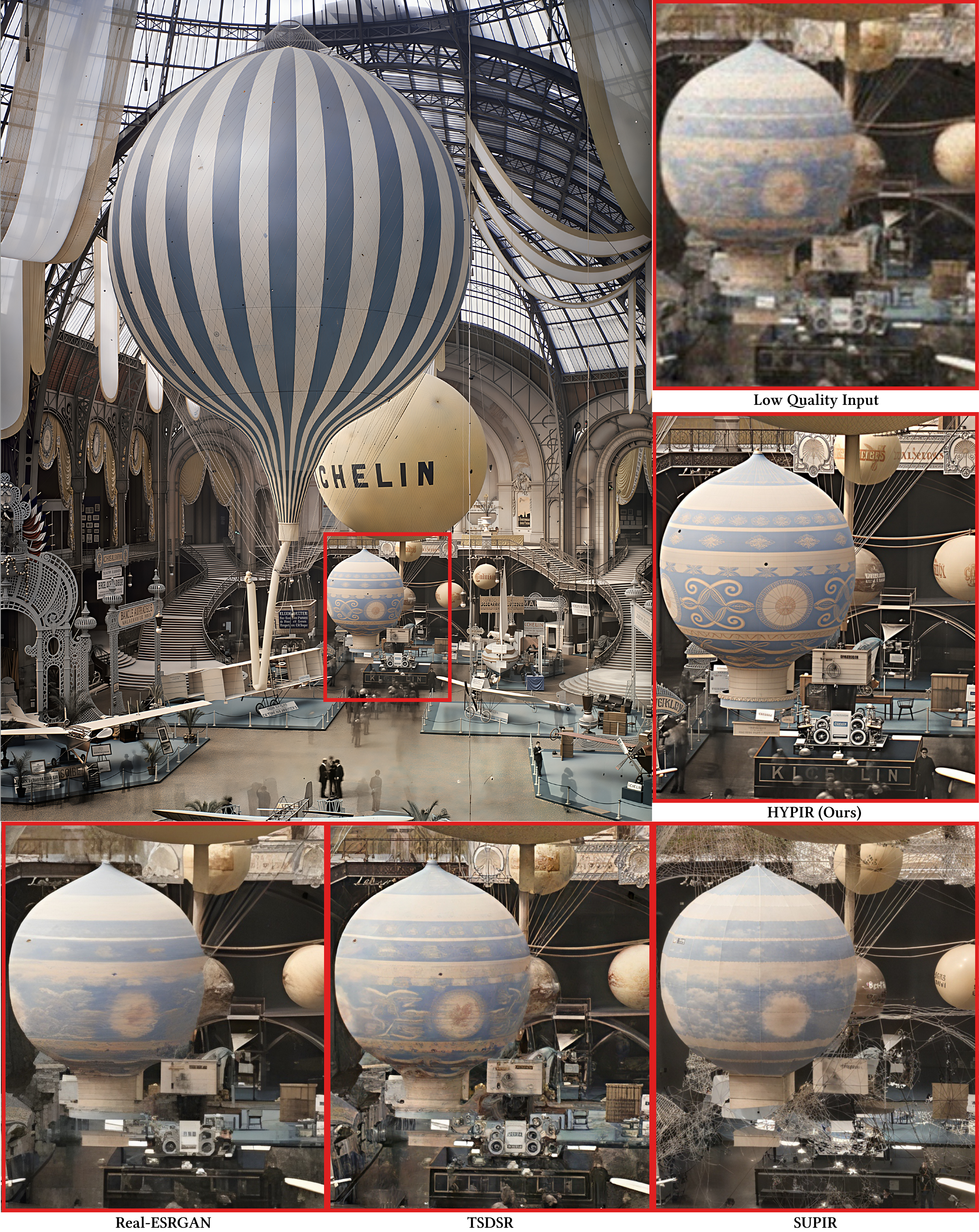}
    \caption{Restoration results for severely degraded historical photographs from over a century ago. Despite low-resolution and complex degradations, HYPIR successfully reconstructs these images at 4K and even 6K resolution, preserving fine details, structural coherence, and authentic textures. Photo Credit: Hot-air balloon, Paris (Autochrome, 1914). Photograph by Auguste \& Louis Lumière, Public Domain. Source: Wikimedia Commons.}
    \Description{Restoration results for severely degraded historical photographs from over a century ago. Despite low-resolution and complex degradations, HYPIR successfully reconstructs these images at 4K and even 6K resolution, preserving fine details, structural coherence, and authentic textures. Photo Credit: Hot-air balloon, Paris (Autochrome, 1914). Photograph by Auguste \& Louis Lumière, Public Domain. Source: Wikimedia Commons.}
    \label{fig:4k}
\end{figure*}

\begin{figure*}
    \centering
    \includegraphics[width=\linewidth]{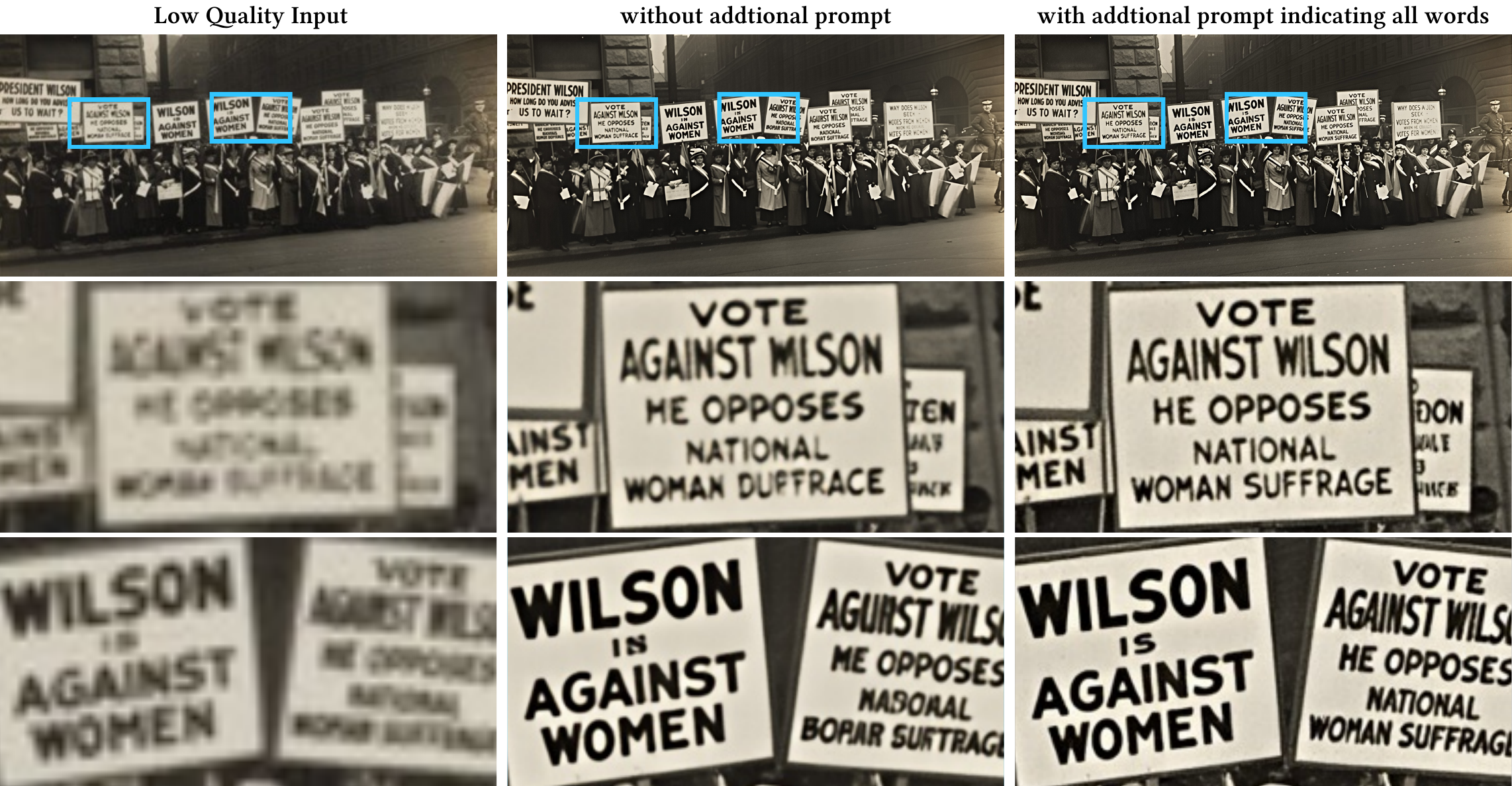}
    \caption{Illustration of text correction capability guided by textual prompts. Without explicit prompts, textual content in degraded images is misinterpreted or inaccurately restored. By providing clear semantic prompts (e.g., ``Vote against Wilson, he opposed national woman suffrage''), HYPIR accurately reconstructs characters, delivering readable, semantically faithful restorations. Photo Credits: Images from the RealLR200 dataset (licensed CC BY 4.0).}
    \Description{Illustration of text correction capability guided by textual prompts. Without explicit prompts, textual content in degraded images is misinterpreted or inaccurately restored. By providing clear semantic prompts (e.g., ``Vote against Wilson, he opposed national woman suffrage''), HYPIR accurately reconstructs characters, delivering readable, semantically faithful restorations. Photo Credits: Images from the RealLR200 dataset (licensed CC BY 4.0).}
    \label{fig:text-ptompt-2}
\end{figure*}

\begin{figure*}
    \centering
    \includegraphics[width=\linewidth]{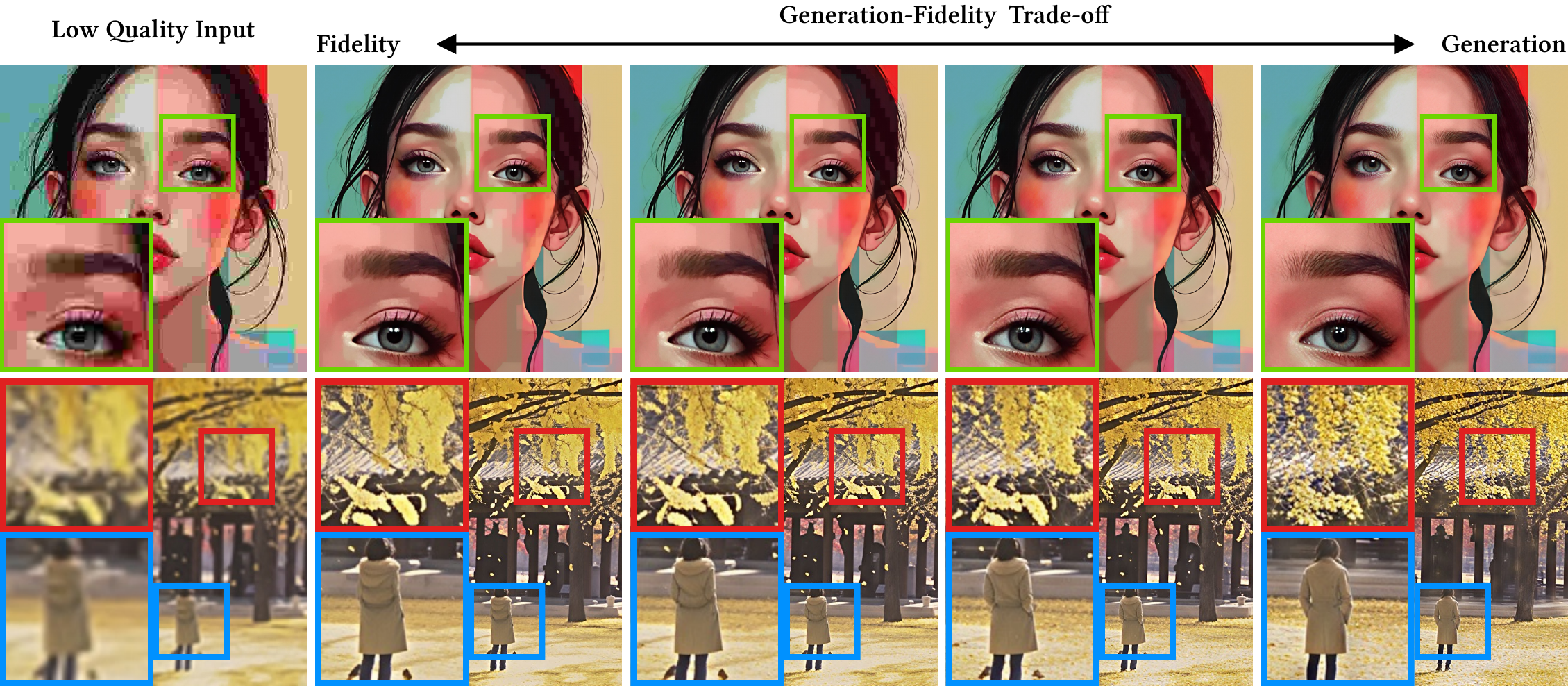}
    \caption{Demonstration of controlling the fidelity-generativity trade-off by introducing artificial noise during restoration. Higher generative ratios help mitigate compression artifacts (first example) and allow the model to synthesize realistic textures for severely blurred regions (second example), improving visual quality when restoring heavily degraded images. Photo Credits: The first image is generated, the second image is from the RealPhoto60 dataset (licensed CC BY 4.0).}
    \Description{Demonstration of controlling the fidelity-generativity trade-off by introducing artificial noise during restoration. Higher generative ratios help mitigate compression artifacts (first example) and allow the model to synthesize realistic textures for severely blurred regions (second example), improving visual quality when restoring heavily degraded images. Photo Credits: The first image is generated, the second image is from the RealPhoto60 dataset (licensed CC BY 4.0).}
    \label{fig:creativity}
\end{figure*}

\begin{figure}
    \centering
    \includegraphics[width=\linewidth]{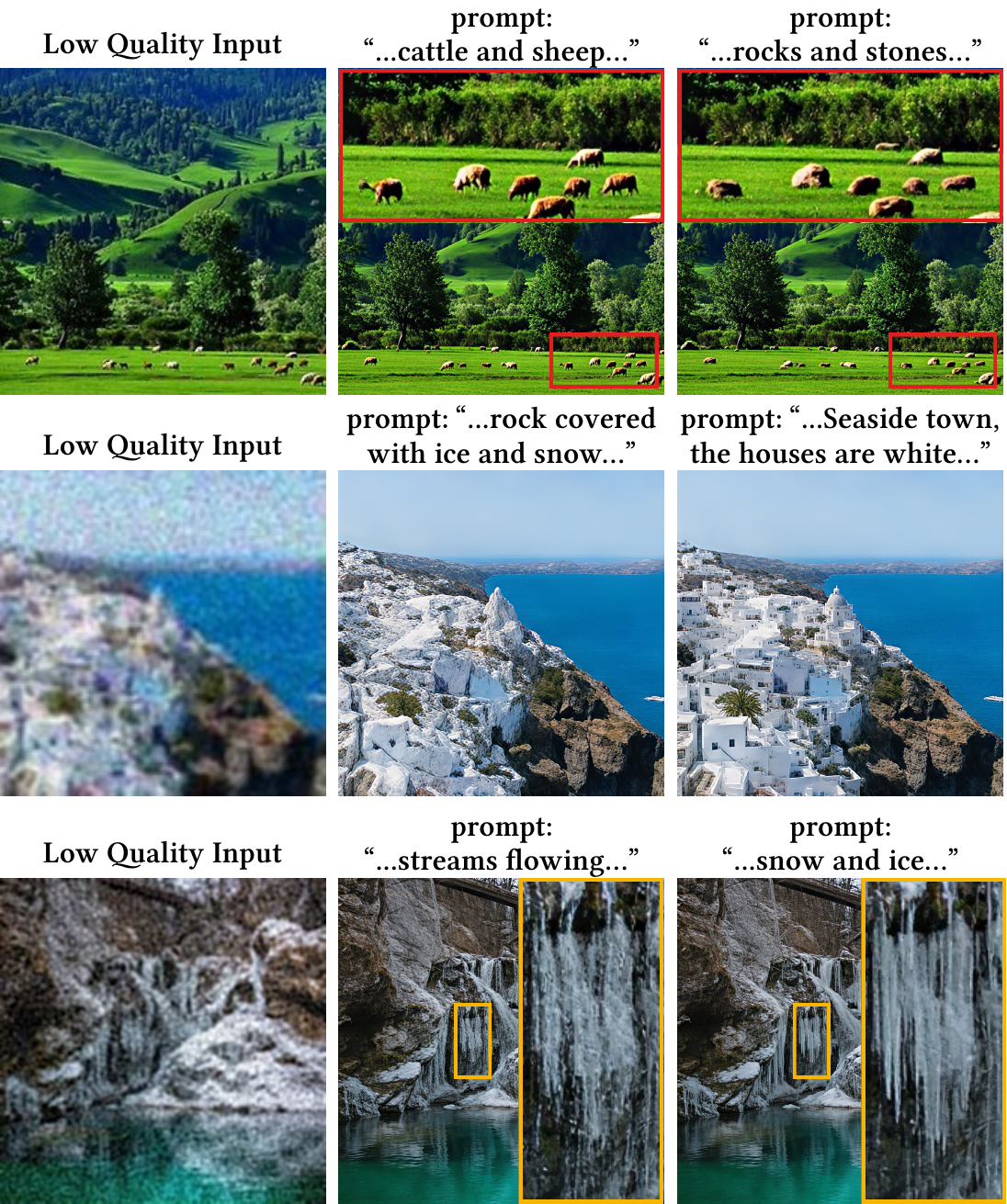}
    \caption{Examples demonstrating semantic guidance of ambiguous image content via textual prompts. Photo Credits: Images from the DIV2K dataset (licensed CC BY 4.0).}
    \Description{Examples demonstrating semantic guidance of ambiguous image content via textual prompts. Photo Credits: Images from the DIV2K dataset (licensed CC BY 4.0).}
    \label{fig:text-ptompt}
\end{figure}

\begin{figure}
    \centering
    \includegraphics[width=\linewidth]{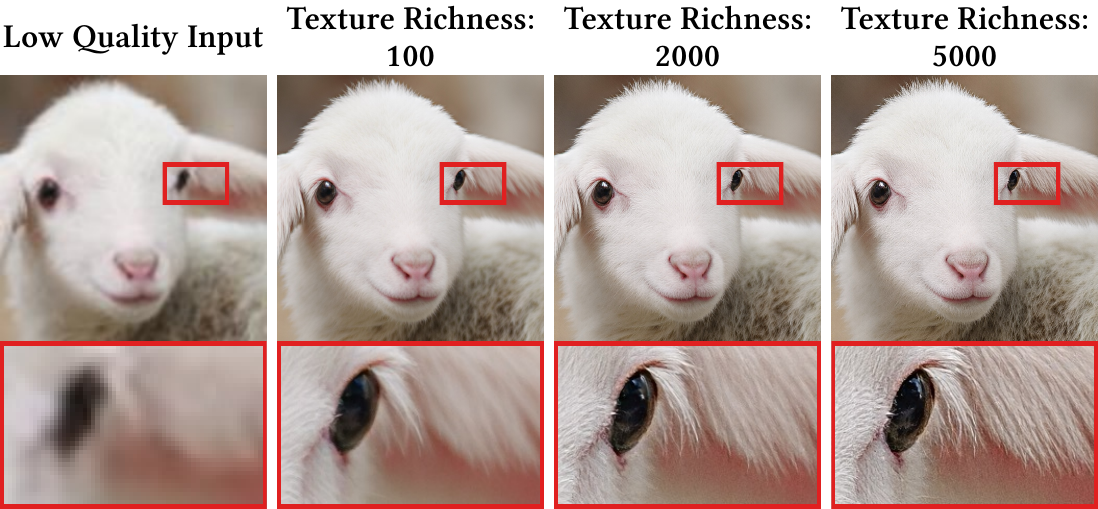}
    \caption{Examples of controlling restored image textures using the proposed texture richness parameter. Lower parameter values yield smoother, softer textures, whereas higher values significantly enhance texture sharpness and detail, enabling intuitive stylistic adjustments. Photo Credits: Images from the RealLR200 dataset (licensed CC BY 4.0).}
    \Description{\textcolor{red}{[The final image is wrong, should be 5000 richness.]} Examples of controlling restored image textures using the proposed texture richness parameter. Lower parameter values yield smoother, softer textures, whereas higher values significantly enhance texture sharpness and detail, enabling intuitive stylistic adjustments. Photo Credits: Images from the RealLR200 dataset (licensed CC BY 4.0).}
    \label{fig:sharpness}
\end{figure}

\begin{figure}
    \centering
    \includegraphics[width=\linewidth]{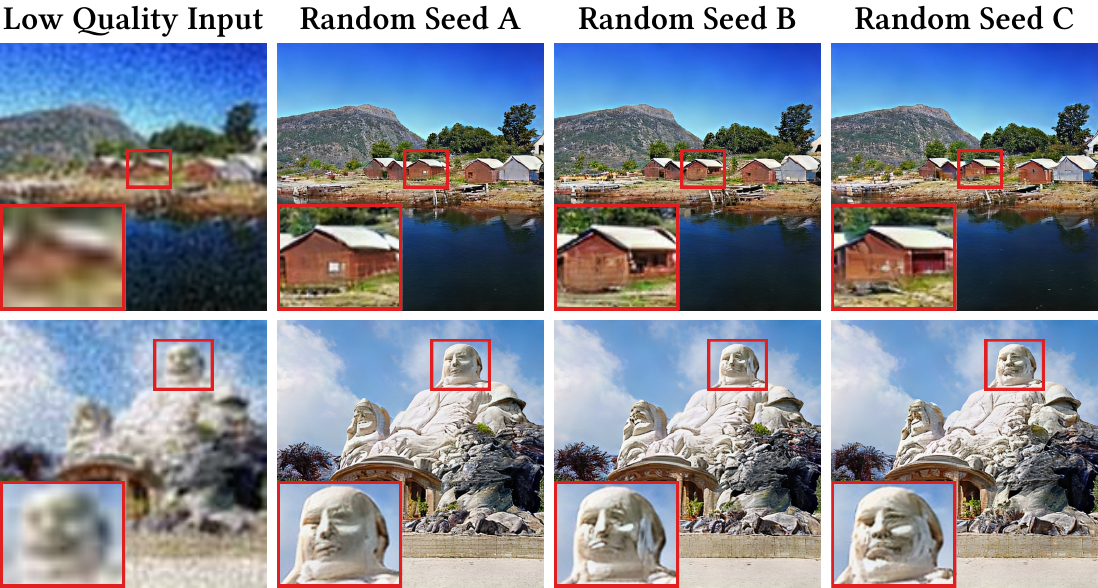}
    \caption{Examples of random sampling by altering concatenated noise inputs. Changing the random seed yields diverse restoration results from the same degraded input, enabling flexible adjustments to local image details. Photo Credits: Images from the DIV2K dataset (licensed CC BY 4.0).}
    \Description{Examples of random sampling by altering concatenated noise inputs. Changing the random seed yields diverse restoration results from the same degraded input, enabling flexible adjustments to local image details. Photo Credits: Images from the DIV2K dataset (licensed CC BY 4.0).}
    \label{fig:seed}
\end{figure}

\subsubsection{Diffusion Priors}
Our approach relies on pretrained diffusion models for initialization.
The size and capability of these diffusion models influence the final performance.
Specifically, larger diffusion models not only provide a more accurate approximation of the score function (see Theorem \ref{thm:diff2rest}), but also exhibit a stronger ability to capture complex image structures.
\figurename~\ref{fig:base-model} illustrates the performance differences when employing various pretrained diffusion models as initialization.
As can be observed, adopting larger and more advanced diffusion models leads to improved restoration outcomes.
It is crucial to note that these performance gains are not merely due to increased model size, without the proposed method, training large-scale GANs alone remains inherently challenging and hard to achieve comparable results.

\subsection{Comparisons}
We conducted comprehensive comparisons between HYPIR and several existing image restoration models.
Specifically, we included the widely adopted classic model Real-ESRGAN \cite{wang2021real}, diffusion-based multi-step methods such as StableSR \cite{wang2024exploiting} and DiffBIR \cite{lin2024diffbir}, the large-scale SDXL-based SUPIR model \cite{yu2024scaling}, as well as distilled or single-step approaches including InvSR \cite{yue2024arbitrary}, TSDSR \cite{dong2024tsd}, S3Diff \cite{zhang2024degradation}, SeeSR \cite{wu2024seesr}, and OSEDiff \cite{wu2024one}.
Evaluations were performed using publicly available datasets DIV2K \cite{div2k} and RealPhoto60 \cite{yu2024scaling}.
Additionally, we introduced a collection of historical photographs to further assess restoration performance on real-world images.
Unless otherwise specified, all results from HYPIR are generated using default parameters, with text prompts automatically produced by the LlaVA model.

\subsubsection{Qualitative Results}
First, we present qualitative comparisons on synthetic test data in \figurename~\ref{fig:comparison}.
Results show that our SD2-based HYPIR method already generates textures superior to the compared approaches, while the Flux-based HYPIR further enhances restoration quality, producing exceptionally clear structural details.
For example, in the first sample of \figurename~\ref{fig:comparison}, the fine-grained structures of garlic and glass bottles are accurately restored only by HYPIR, whereas other methods yield blurry and indistinct results.
\figurename~\ref{fig:real-comparison} demonstrates restoration results on a variety of images collected from real-world scenes.
Once again, our method shows excellent performance in recovering intricate details such as facial features, textual content, and architectural structures.
Additionally, \figurename~\ref{fig:4k} and \figurename~\ref{fig:teaser} illustrate the capability of HYPIR in restoring historical photographs taken over a century ago.
Despite severe degradation and low-resolution inputs, HYPIR effectively reconstructs these images at resolutions up to 4K or even 6K, maintaining impressive fidelity in local details, structural integrity, and textures.

\subsubsection{Quantitative Evaluation}
To quantitatively evaluate the performance of HYPIR, we conducted both user studies and analyses based on Image Quality Assessment (IQA) metrics.
Note that IQA metrics are inadequate for evaluating results from generative-model-based image restoration methods, as reported by many studies \cite{jinjin2020pipal,yu2024scaling}.
Specifically, it is possible to artificially boost scores on these metrics without achieving any improvement in perceived visual quality.
Thus, IQA metrics are presented here only for reference purposes, and we place greater emphasis on the user study, which directly reflects perceptual quality.

We conducted two rounds of user studies involving 26 images sampled from our test dataset (16 real images and 10 synthetic images).
In the first round (\figurename~\ref{fig:user_round1}), we compared models that are developed from lightweight base models like SD2 (0.8B), SD-turbo (0.8B), or SD3-medium (2B). 
In the second round (\figurename~\ref{fig:user_round2}), we expanded our comparisons to include the SUPIR model based on SDXL (2.6B), as well as HYPIR based on Flux (12B).

A total of 100 participants were recruited through Prolific \cite{prolific} to rate each restored image.
We required participants to have prior experience in image processing, complete the survey using a desktop computer, and view the full-sized images on large screens.
Participants rated each image on an 11-level scale from 0 to 10, where higher scores indicate better visual preference.
Participants were encouraged to maintain monotonic consistency in their ratings, ensuring that better-quality images received higher scores relative to poorer ones.
The results of the user studies are presented in \figurename~\ref{fig:user_round1} and \figurename~\ref{fig:user_round2}.
As can be seen, HYPIR achieves the highest perceptual scores in both evaluation rounds.
Furthermore, instances of low ratings for HYPIR are infrequent, demonstrating the robustness and reliability of our approach.

In Table~\ref{tab:res_quantitative}, we report quantitative results on both synthetic image and natural image datasets using non-reference image quality metrics, including NIQE~\cite{niqe}, MUSIQ~\cite{musiq}, MANIQA~\cite{maniqa}, CLIP-IQA~\cite{clipiqa}, and DeQA-Score~\cite{deqa}. 
For the synthetic dataset, where ground-truth images are available, we additionally report full-reference quality metrics such as PSNR, SSIM~\cite{ssim}, and LPIPS~\cite{lpips}.
As shown in Table~\ref{tab:res_quantitative}, our method consistently ranks among the top-performing methods according to these metrics.
However, we reiterate that these quantitative metrics alone do not fully capture human perceptual quality.
Therefore, while we present these objective metrics for completeness, our primary focus remains on perceptual quality as reflected by user evaluations.

\subsubsection{Running Time}
We further compare the inference speed of various restoration methods.
\figurename~\ref{fig:time} shows the average inference time required by different algorithms to produce images of size $1024\times1024$ (averaged over 100 test images) on a Nvidia A6000 GPU.
It can be observed that multi-step diffusion models require significantly more inference time due to their iterative sampling procedure.
Conversely, single-step models achieve faster inference but often compromise on restoration quality. Remarkably, our approach, even when employing large-scale models such as Flux (12B parameters), achieves exceptional restoration quality while maintaining inference speeds comparable to other single-step methods.

\subsection{Conditional Image Restoration}
In Section \ref{sec:control}, we described four strategies to control image restoration details during inference.
Here, we present empirical demonstrations of these strategies in practice.

\subsubsection{Control by Text Prompt}
By inheriting the architecture and parameters of the pre-trained diffusion model, HYPIR naturally acquires the ability to understand and respond to textual prompts.
This allows users to interact with the restoration model via natural language instructions, guiding the restoration process according to their specific intentions.
Our design supports controlled, semantically aware restoration.
We show the following examples.

\paragraph{Correcting textual elements}
Texts within degraded images, such as signs or documents, often become severely distorted due to noise, blur, or other degradations.
Recovering the original textual content solely from damaged visual clues (the left image of \figurename~\ref{fig:text-ptompt-2}) is challenging, as degraded images typically lack sufficient information for accurately inferring the intended characters.
Our model leverages the strong textual prior provided by the text-to-image diffusion model, enabling interpretation and synthesis of textual content based on natural language descriptions.
This functionality allows users to provide additional text prompts explicitly specifying the desired content in degraded text regions.
As shown in \figurename~\ref{fig:text-ptompt-2}, prompts such as ``Vote against Wilson, he opposed national woman suffrage'' or ``…Wilson is against…'' clearly convey semantic intent.
With such prompts, the model can reconstruct clear, readable, and semantically faithful characters, whereas failing to provide these prompts leads to incorrect results.

\paragraph{Identifying and refining ambiguous objects}
Image restoration is an inherently ill-posed problem, where a single degraded input can correspond to multiple plausible solutions.
Without additional constraints, models often struggle to generate desired outputs.
Introducing textual prompts helps narrow the solution space and guides the model toward a specific generation direction.
In \figurename~\ref{fig:text-ptompt}, the first case illustrates cases where certain objects occupy too few pixels and appear highly ambiguous due to poor input image quality.
By providing different prompts (e.g., ``…cows and sheep…'' or ``…rocks and stones…''), users can guide the model toward generating customized, semantically coherent content.
The second case demonstrates that textual prompts can influence not only small or localized objects but also the overall image structure.
Even when image quality is extremely poor and the scene is barely recognizable by human observers, our model can faithfully follow user instructions to produce intended outcomes.
The third case highlights the model's capability to distinguish between semantically similar concepts.
Prompts such as ``…streams…'' or ``…snow and ice…'' enable fine-grained control over object properties, allowing users to precisely manipulate subtle aspects of the restored images.

Overall, HYPIR excels in accurately interpreting textual prompts, especially in reconstructing text characters with high fidelity -- a capability that significantly surpasses limitations in previous state-of-the-art methods \cite{yu2024scaling}.
Our interactive framework thus enables users to achieve personalized, meaningful, and high-quality restoration outcomes.

\subsubsection{Control by Texture Richness}
\figurename~\ref{fig:sharpness} illustrates an example demonstrating the controllability of restoration using the texture richness parameter.
It can be observed that lower values of this parameter yield smoother and softer textures, particularly noticeable in regions such as animal fur.
Conversely, increasing the texture richness parameter significantly enhances sharpness and detail in the fur textures.
This functionality provides users with intuitive control over the overall stylistic appearance of restored images.

\subsubsection{Control by Generativity-Fidelity Trade-off}
We further achieve a controllable trade-off between fidelity and generative quality by introducing artificial noise during the restoration process.
This can effectively obscure certain original details and guide the model to reconstruct these areas creatively.
\figurename~\ref{fig:creativity} illustrates two representative examples.
In the first example, the input image exhibits severe compression artifacts, resulting in noticeable grid-like patterns upon restoration. 
While increasing the generative ratio -- \textit{i.e.,} intentionally adding noise -- we encourage the model to produce smoother textures, effectively removing the undesired grid patterns.
Similarly, in the second example, when the generative ratio is low, the model struggles to associate severely blurred regions with meaningful textures.
Increasing this ratio allows the model to transcend the constraints imposed by the degraded input, synthesizing realistic and detailed textures.
Such controllability is particularly beneficial for handling images with severe degradations.

\subsubsection{Control by Random Sampling}
Finally, we illustrate how results can be randomly sampled by adjusting the concatenated noise input.
Such stochastic sampling allows users to explore alternative restoration outcomes, particularly useful when local regions are unsatisfactory in the initial results.
\figurename~\ref{fig:seed} demonstrates how altering the random seed produces diverse restoration outputs from the same degraded image, highlighting the flexibility of our approach in generating multiple plausible restorations.

\section{Conclusion}
In this work, we introduced \textbf{HYPIR}, a novel yet remarkably straightforward image restoration framework.
By harnessing the powerful prior encapsulated within pretrained diffusion models for initialization, and employing lightweight adversarial fine-tuning via LoRA, our approach achieves rapid convergence, numerical stability, and outstanding restoration quality.
Unlike conventional diffusion-based methods, HYPIR completely eliminates iterative sampling, distillation complexity, and computationally demanding auxiliary networks.
Consequently, it significantly accelerates both training and inference speed without sacrificing restoration performance -- indeed, it consistently surpasses state-of-the-art methods.
Moreover, HYPIR uniquely empowers users with intuitive control mechanisms such as textual prompts, adjustable texture richness, and customizable fidelity-generativity balance.



\bibliographystyle{ACM-Reference-Format}
\bibliography{sample-base}


\appendix

\section*{Appendix}

\section{Proofs}

\subsection{Proof of Theorem~\ref{thm:diff2rest}}
\label{thm:diff2rest:proof}
\paragraph{Step 1 (diffusion accuracy).}
Let \(\tilde p:=p_{\mathrm{data}} *k_{\sigma}\) and write
\(
  \widetilde{p}_{\theta_{\mathrm{Diff}}}
  :=\mathcal U_{\theta_{\mathrm{Diff}}} \sharp\,\tilde p.
\)
\citet{song2019generative} probability-flow ODE result yields
\[W_{2}(\widetilde{p}_{\theta_{D}},p_{\mathrm{data}})\le C_{1}\,\varepsilon_{\mathrm{sc}},\]
where
\[
\varepsilon_{\mathrm{sc}}
\;:=\;
\Bigl(
  \int
  \bigl\|
     \mathcal S_{\theta_{\mathrm{Diff}}}(x,\sigma)
     -\nabla_{x}\log(p_{\mathrm{data}} *k_{\sigma})(x)
  \bigr\|^{2}
  p_{\mathrm{data}}(x)\,dx
\Bigr)^{1/2}
\]

\paragraph{Step 2 (kernel mismatch).}
Young’s convolutional inequality gives
\begin{align}
    \lVert p_{y}-\tilde p\rVert_{1}&=\bigl\|p_y-(p_{\textup{data}} *k_\sigma)\bigr\|_{1}\\&=
   \bigl\|p_{\textup{data}} * (k_{\textup{deg}}-k_\sigma)\bigr\|_{1}
   \\&\le
   \|p_{\textup{data}}\|_{1}\,\|k_{\textup{deg}}-k_\sigma\|_{1}\\&=\Delta_k.
\end{align}

By Pinsker and Kantorovich–Rubinstein,
\(
  W_{2} \bigl(p_{\theta_{\mathrm{Diff}}},\widetilde{p}_{\theta_{\mathrm{Diff}}}
  \bigr)
  \le
  C_{2}\,\Delta_{k},
\)
where \(C_{2}\) absorbs the Lipschitz bound of
\(\mathcal U_{\theta_{\mathrm{Diff}}}\).

\paragraph{Step 3 (triangle inequality).}
Finally,
\[
  W_{2}(p_{\theta_{\mathrm{Diff}}},p_{\mathrm{data}})
  \le
  W_{2}(p_{\theta_{\mathrm{Diff}}},\widetilde{p}_{\theta_{\mathrm{Diff}}})
  +
  W_{2}(\widetilde{p}_{\theta_{\mathrm{Diff}}},p_{\mathrm{data}})
  \le
  C_{1}\,\varepsilon_{\mathrm{sc}}
  +
  C_{2}\,\Delta_{k},
\]
which is the claimed bound.

\hfill$\square$

\subsection{Proof of Lemma~\ref{lem:init-grad}}
\label{lem:init-grad:proof}
With $D^\star_\theta$ the optimal discriminator
(\citet{arjovsky2017towards}),  
\[
   \mathcal L_G(\theta)
   \;=\;
   -\mathbb E_{y\sim p_y} \Bigl[
        \log\frac{p_{\mathrm{data}}}
                  {p_{\mathrm{data}}+p_\theta}
            \Bigl(\mathcal R_\theta(y)\Bigr)
   \Bigr].
\]
Differentiating under the integral\footnote{Justified because (i) the integrand is $C^1$ in $\theta$ by bounded Jacobian, and (ii) dominated convergence holds since the log–density ratio is bounded by $\log 2$.} and applying the chain rule yields 
\begin{equation}
   \nabla_\theta\mathcal L_G(\theta)
   \;=\;
   \mathbb E_{y\sim p_y} \Bigl[
        J_\theta(y)
        \,\frac{p_{\mathrm{data}}-p_\theta}
               {p_{\mathrm{data}}}
        \Bigl(x\Bigr)
        \Bigl|_{x=\mathcal R_\theta(y)}
   \Bigr].
   \label{eq:S1-fraction}
\end{equation}

Take operator norm and apply Cauchy–Schwarz for the integration measure $\mu(dy)=p_y(y)\,dy$:
\[
   \bigl\|\nabla_\theta\mathcal L_G(\theta) \bigr\|
    \le 
   \Bigl(
     \mathbb E_y\|J_\theta(y)\|^2
   \Bigr)^{ 1/2} 
   \Bigl(
     \mathbb E_y
     \Bigl|
       \frac{p_{\mathrm{data}} - p_\theta}{p_{\mathrm{data}}}
         \bigl(\mathcal R_\theta(y)\bigr)
     \Bigr|^2
   \Bigr)^{ 1/2}.
   \tag{S2-a}
\]
Using (b) gives the first factor $\le L_J$. For the second factor, we change variables $x=\mathcal R_\theta(y)$, whose Jacobian determinant we bound by $|\det\nabla_y\mathcal R_\theta|\le L_g^{d_x}$ via (a).
Thus
\[
   \mathbb E_y\Bigl|
       \tfrac{p_{\mathrm{data}}-p_\theta}{p_{\mathrm{data}}}
         \bigl(\mathcal R_\theta(y)\bigr)
   \Bigr|^2
   \le
   L_g^{d_x}
   \int_{\mathbb R^{d_x}}
     \frac{|p_{\mathrm{data}}(x)-p_\theta(x)|^2}
          {p_{\mathrm{data}}(x)}
   \,dx.
   \tag{S2-b}
\]
Since $p_{\mathrm{data}}\ge0$ and $\le M$,  
\(|p_{\mathrm{data}}-p_\theta|^2\le M|p_{\mathrm{data}}-p_\theta|\).
Hence the integral upper–bounds by
$M\|p_{\mathrm{data}}-p_\theta\|_1$.

From $L^1$ to $W_2$, Pinsker’s inequality gives  
\(
   \|p_{\mathrm{data}}-p_\theta\|_1
   \le\sqrt2\,W_2(p_{\mathrm{data}},p_\theta).
\)
Evaluating at $\theta=\theta_D$ and
invoking Theorem~\ref{thm:diff2rest} yields
$W_2\le\epsilon_0$.

Combine (S2-a)–(S2-b):
\[
   \|\nabla_\theta\mathcal L_G(\theta_D)\|
   \;\le\;
   L_J\,
   \bigl(L_g^{d_x}M\bigr)^{1/2}\,
   \sqrt{\sqrt2\,\epsilon_0}.
\]
Absorb $\bigl(L_g^{d_x}M\bigr)^{1/2}$ into $L_J$
(by redefining $L_J\gets L_J\sqrt{L_g^{d_x}M}$; one may also keep it explicit).
Renaming constant gives the desired
$\sqrt2\,L_J\,\epsilon_0$ bound.

\hfill$\square$

\subsection{Proof of Proposition~\ref{prop:mode}}
\label{prop:mode:proof}

\paragraph{Step 1 (Writing the probability mass difference as an integral).}
Fix a region\footnote{It does \emph{not} matter how the regions~$\mathcal A_k$ are chosen: they can be semantic classes (``cats'', ``dogs''), clusters in feature space, or even arbitrary disjoint measurable sets.} $\mathcal A_k\subset\mathbb R^{d_x}$.
By definition of probability,
\[
   p_{\theta_D}(\mathcal A_k)
   \;=\;\int_{\mathcal A_k} p_{\theta_D}(x)\,dx,
   \quad
   p_{\mathrm{data}}(\mathcal A_k)
   \;=\;\int_{\mathcal A_k} p_{\mathrm{data}}(x)\,dx.
\]
Therefore
\begin{equation}
      \bigl|
    p_{\theta_D}(\mathcal A_k)-p_{\mathrm{data}}(\mathcal A_k)
  \bigr|
  \;=\;
  \Bigl|
    \int_{\mathcal A_k}
      \bigl[p_{\theta_D}(x)-p_{\mathrm{data}}(x)\bigr]\,dx
  \Bigr|.
\label{eq:S1}
\end{equation}

\paragraph{Step 2 (Bounding the integral by the total variation
distance).}
Introduce the indicator function
\(
  \mathbf 1_{\mathcal A_k}(x)=1
\)
if $x\in\mathcal A_k$ and $0$ otherwise. Then~Equation~\eqref{eq:S1} reads
\[
   \Bigl|\int
          \mathbf 1_{\mathcal A_k}(x)
          \bigl[p_{\theta_D}(x)-p_{\mathrm{data}}(x)\bigr]\,
          dx\Bigr|.
\]
For \emph{any} integrable function $f(x)$ whose values lie between $-1$ and $1$ (such as an indicator taking only $0/1$), Hölder’s inequality\footnote{Hölder in 1-D is the familiar Cauchy–Schwarz
with absolute values.} states
\(
   | \int f(x)\,g(x)\,dx|
   \le
   \int |g(x)|\,dx.
\)
Taking $g(x)=p_{\theta_D}(x)-p_{\mathrm{data}}(x)$ and $f=\mathbf 1_{\mathcal A_k}$ yields
\[
   \bigl|
     p_{\theta_D}(\mathcal A_k)-p_{\mathrm{data}}(\mathcal A_k)
   \bigr|
   \;\le\;
   \tfrac12 \int
     |p_{\theta_D}(x)-p_{\mathrm{data}}(x)|\,dx,
\]
where the factor $\tfrac12$ is introduced by convention so that the right hand side becomes the total variation distance
\[
    \mathrm{TV}(p_{\theta_D},p_{\mathrm{data}})
    \;=\;
    \tfrac12\,\|p_{\theta_D}-p_{\mathrm{data}}\|_{L^1}.
\]
Hence
\begin{equation}
       \bigl|
     p_{\theta_D}(\mathcal A_k)-p_{\mathrm{data}}(\mathcal A_k)
   \bigr|
   \;\le\;
   \mathrm{TV}(p_{\theta_D},p_{\mathrm{data}}).
\label{eq:S2}
\end{equation}

\paragraph{Step 3 (Linking total variation to the 2-Wasserstein distance).}
A classical inequality states that
\begin{equation}
       \mathrm{TV}(p,q)
   \;\le\;
   \sqrt2\,W_2(p,q).
\label{eq:S3}
\end{equation}
Intuitively, if two distributions are close in the $W_2$ sense, we never have to move much probability mass very far, so the amount of mass that needs to be moved (i.e.\ the total variation) must also be small.

\paragraph{Step 4 (Using Theorem~\ref{thm:diff2rest}).}
From Theorem~\ref{thm:diff2rest} we already know that \(W_2(p_{\theta_D},p_{\mathrm{data}})\le\epsilon_0\).
Plugging this and~Equation~\eqref{eq:S3} into~Equation~\eqref{eq:S2} yields
\[
   \bigl|
     p_{\theta_D}(\mathcal A_k)-p_{\mathrm{data}}(\mathcal A_k)
   \bigr|
   \;\le\;
   \sqrt2\,\epsilon_0
   \;=\;
   \frac{\sqrt2}{2}\,(2\epsilon_0),
\]
and replacing the innocuous factor $2$ inside the parenthesis by the
constant finishes the proof.

\hfill$\square$

\subsection{Proof of Proposition~\ref{prop:linlog}}
\label{prop:linlog:proof}

We prove that gradient descent needs only \(\mathcal O(\log(1/\epsilon_0))\) iterations to shrink the loss gap below a target \(\delta_{\mathrm{tar}}\).
The argument has three ingredients:
\begin{enumerate}[label=(\arabic*)]
\item a \emph{smoothness lemma} that tells us how much the loss can drop in one step;
\item a \emph{strong-convexity lemma} that links loss gap to
      parameter distance;
\item a short calculation that chains the two lemmas to count steps.
\end{enumerate}

\paragraph{(1) Smoothness lemma.}
Because \(\mathcal L_G\) is \(L\)-smooth, moving one step by \(-\eta\nabla\mathcal L_G(\theta^t)\) cannot increase the loss too quickly
\[
   \mathcal L_G(\theta^{t+1})
   \le
   \mathcal L_G(\theta^{t})
   -\Bigl(\eta-\tfrac{L\eta^{2}}{2}\Bigr)
       \,\bigl\|\nabla\mathcal L_G(\theta^{t})\bigr\|^{2}.
\]
Choosing \(\eta\le 1/L\) makes the bracket
\((\eta-L\eta^{2}/2)\ge\eta/2\), so
\[
   \mathcal L_G(\theta^{t+1})
   \le
   \mathcal L_G(\theta^{t})
   -\frac{\eta}{2}\,
       \bigl\|\nabla\mathcal L_G(\theta^{t})\bigr\|^{2}.
\tag{A.1}
\]

\paragraph{(2) Strong-convexity lemma.}
Under \(\mu\)-strong convexity, the current loss gap \(\delta_t\) provides a lower bound on the squared gradient norm:
\[
   \bigl\|\nabla\mathcal L_G(\theta^{t})\bigr\|^{2}
   \;\ge\;
   2\mu\,\delta_t.
\tag{A.2}
\]

Because $\mathcal L_G$ is $\mu$-strongly convex, taking $\theta_1=\theta^{t}$ and the minimizer $\theta_2=\theta^\star$ in the definition
$\mathcal L_G(\theta_2)\ge\mathcal L_G(\theta_1)+\langle\nabla\mathcal L_G(\theta_1),\theta_2-\theta_1\rangle+\tfrac{\mu}{2}\|\theta_2-\theta_1\|^{2}$ yields the lower bound $\tfrac{\mu}{2}\|\theta^{t}-\theta^\star\|^{2}\le\delta_t$.
Applying the same inequality but rearranging the inner product and then invoking Cauchy–Schwarz gives $\langle\nabla\mathcal L_G(\theta^{t}),\theta^{t}-\theta^\star\rangle\ge\mu\|\theta^{t}-\theta^\star\|^{2}$, and dividing both sides by $\|\theta^{t}-\theta^\star\|$ provides the gradient estimate $\|\nabla\mathcal L_G(\theta^{t})\|\ge\mu\|\theta^{t}-\theta^\star\|$.

\paragraph{(3) Chaining the lemmas (linear decay).}
Insert (A.2) into (A.1):
\[
   \delta_{t+1}
   \;\le\;
   \delta_t
   -\eta\mu\,\delta_t
   \;=\;
   (1-\eta\mu)\,\delta_t.
\]
Thus the loss gap shrinks by a constant factor
\(q:=1-\eta\mu<1\) \emph{every iteration}:
\[
   \delta_t
   \;\le\;
   q^{\,t}\,\delta_0.
\tag{A.3}
\]

\paragraph{(4) Linking \(\delta_0\) to \(\epsilon_0\).}
\citet{gulrajani2017improved} show
\[\delta_0\le C_2\,W_2^{2}(p_{\theta_D},p_{\mathrm{data}})\le C_2\,\epsilon_0^{2}.\]
So (A.3) becomes
\[
   \delta_t
   \;\le\;
   C_2\,\epsilon_0^{2}\;q^{\,t}.
\tag{A.4}
\]

\paragraph{(5) Counting steps.}
We want \(\delta_t\le\delta_{\mathrm{tar}}\).  
Solve (A.4) for \(t\):
\[
 q^{\,t}\le \frac{\delta_{\mathrm{tar}}}{C_2\epsilon_0^{2}}
 \quad\Longrightarrow\quad
 t\;\ge\;\frac{\ln\bigl(\tfrac{C_2\epsilon_0^{2}}{\delta_{\mathrm{tar}}}\bigr)}
               {\ln(1/q)}
       \;=\;
       \frac{\ln\bigl(\tfrac{C_2\epsilon_0^{2}}{\delta_{\mathrm{tar}}}\bigr)}
            {\ln\bigl(\tfrac{1}{1-\eta\mu}\bigr)}.
\]
Because the denominator \(\ln(1/(1-\eta\mu))\approx\eta\mu\) is a constant,
the dominating term in the numerator is
\(\ln(\epsilon_0^{2})\), i.e.\ \(\mathcal O(\log\tfrac1{\epsilon_0})\).

\hfill$\square$

\end{document}